\documentclass[10pt,journal,compsoc]{IEEEtran}
\usepackage{cite}
\usepackage{times}
\usepackage{epsfig}
\usepackage{graphicx}
\usepackage{amsmath}
\usepackage{amssymb}
\usepackage{multirow}
\usepackage{longtable}
\usepackage{color}
\usepackage{rotating}
\usepackage{makecell}
\usepackage{array}
\usepackage{booktabs}
\usepackage{colortbl}
\usepackage{arydshln}

\usepackage{amsmath}        
\usepackage{graphicx}
\usepackage{multirow}
\usepackage{rotating}      
\usepackage{arydshln}		
\usepackage{wrapfig}       
\usepackage{array}         
\usepackage{bm}            
\usepackage{graphicx}
\usepackage{subfigure}
\usepackage{color}         
\usepackage{enumitem}
\usepackage{colortbl}
\usepackage{indentfirst}   
\usepackage{url}		

\ifCLASSINFOpdf
\else
\fi
\begin{document}
%
\title{Distribution Knowledge Embedding for Graph Pooling}
%
%
%
%

\author{Kaixuan~Chen,
        Jie~Song,
	   Shunyu~Liu,
	   Na~Yu,
	   Zunlei~Feng,
	   Gengshi~Han,
        and~Mingli~Song*,~\IEEEmembership{Senior~Member,~IEEE}
\IEEEcompsocitemizethanks{\IEEEcompsocthanksitem K.~Chen,  J.~Song, S.~Liu, N.~Yu, Z.~Feng, G.~Han, and M.~Song  are with the College of Computer Science and Technology, Zhejiang University, Hangzhou 310027, China.\protect\\
E-mail: chenkx, sjie, liushunyu, na\_yu, zunleifeng, hangengshi, brooksong @zju.edu.cn
\IEEEcompsocthanksitem M.~Song is the corresponding author. E-mail: brooksong@zju.edu.cn}
\thanks{Manuscript received April 19, 2005; revised August 26, 2015.}}

\markboth{Journal of \LaTeX\ Class Files,~Vol.~14, No.~8, August~2015}%
{Shell \MakeLowercase{\textit{et al.}}: Bare Demo of IEEEtran.cls for Computer Society Journals}

\IEEEtitleabstractindextext{%
\begin{abstract}
Graph-level representation learning is the pivotal step for downstream tasks that operate on the whole graph. The most common approach to this problem is graph pooling, where node features are typically averaged or summed to obtain the graph representations.
However, pooling operations like averaging or summing inevitably cause severe information missing, which may severely downgrade the final performance. In this paper, we argue what is crucial to graph-level downstream tasks includes not only the topological structure but also the \textit{distribution} from which nodes are sampled. Therefore, powered by existing Graph Neural Networks~(GNN), we propose a new plug-and-play pooling module, termed as \textit{Distribution Knowledge Embedding}~(DKEPool), where graphs are viewed as distributions on top of GNNs and the pooling goal is to summarize the entire distribution information instead of retaining a certain feature vector by simple predefined pooling operations. A DKEPool network \textit{de facto} disassembles representation learning into two stages, \textit{structure learning} and \textit{distribution learning}. Structure learning follows a recursive neighborhood aggregation scheme to update node features where structure information is obtained. Distribution learning, on the other hand, omits node interconnections and focuses more on the distribution depicted by all the nodes. 
Extensive experiments on graph classification tasks demonstrate that the proposed DKEPool significantly and consistently outperforms the state-of-the-art methods.\\
The code is avaliable at~\url{https://github.com/chenchkx/dkepool}
\end{abstract}

}

\maketitle

\IEEEdisplaynontitleabstractindextext

%
\IEEEpeerreviewmaketitle

\IEEEraisesectionheading{\section{Introduction}\label{sec:introduction}}
\IEEEPARstart{R}{emarkable} successes have been achieved recently by generalizing deep neural networks from grid-like data to graph-structure data, resulting in the rapid development of graph neural networks (GNNs)~\cite{yang2020factorizable,kipf2016semi,xu2018powerful}. GNNs are amenable to learning representations for each instance in the graph and have established new performance records on instance-level tasks like node classification~\cite{hao2021walking} and link prediction~\cite{cai2020multi}. However, when it comes to graph-level tasks, the capacity of GNNs for graph-level representations is challenged by two peculiarities of graphs. First, there is no fixed ordering relationship among graph nodes, which requires that the graph pooling operation should be invariant to the node order.
Second, the number of nodes in different graphs is inconsistent, while a common requirement for most machine learning methods is that the input sample representation should be the same size. 

Graph pooling is required to extract the powerful graph-level representations for the task of graph-level predictions. Existing graph pooling methods can be broadly categorized into two schools, \textit{hierarchical} graph pooling~\cite{ying2018hierarchical,BianchiGA20,ning2021HAP,yang2020hierarchical} and \textit{flat} graph pooling~\cite{duvenaud2015convolutional,defferrard2016convolutional,wang2020second}. Hierarchical graph pooling iteratively operates on coarser and coarser representations of a graph, which involves large network modifications and can hardly be applied to off-the-shelf GNNs. Flat graph pooling, on the other hand, first generates embeddings for all the nodes in the graph and then globally pools all these node embeddings in one step, ignoring the non-Euclidean geometry information of graphs. Both pooling methods suffer from severe information missing, which hampers the settlement of graph-level tasks.

The graphs are non-Euclidean geometry data containing massive undecided non-linear information. In this paper, we argue \textit{what is crucial to graph-level downstream tasks includes not only the topological structure but also the underlying distribution information defined over node features}. 
Motivated by this observation, we propose a new pooling module, termed as \textit{Distribution Knowledge Embedding}~(DKEPool), where graphs are viewed as distributions and the pooling goal is to summarize the entire distribution information instead of retaining a certain feature vector by simple predefined pooling operations. Notably, a distribution depicted by its sampled nodes is also insensitive to the size and order of these nodes, which ensures that DKEPool can output the fixed-sized graph representations when the size of input graphs is different and the same representation when the order of nodes of an input graph changes. Moreover, distribution is endowed with a larger capacity, rendering it a more robust representation for graphs than the simple aggregated feature vector.

\begin{figure*}
\centering
\includegraphics[scale=0.61]{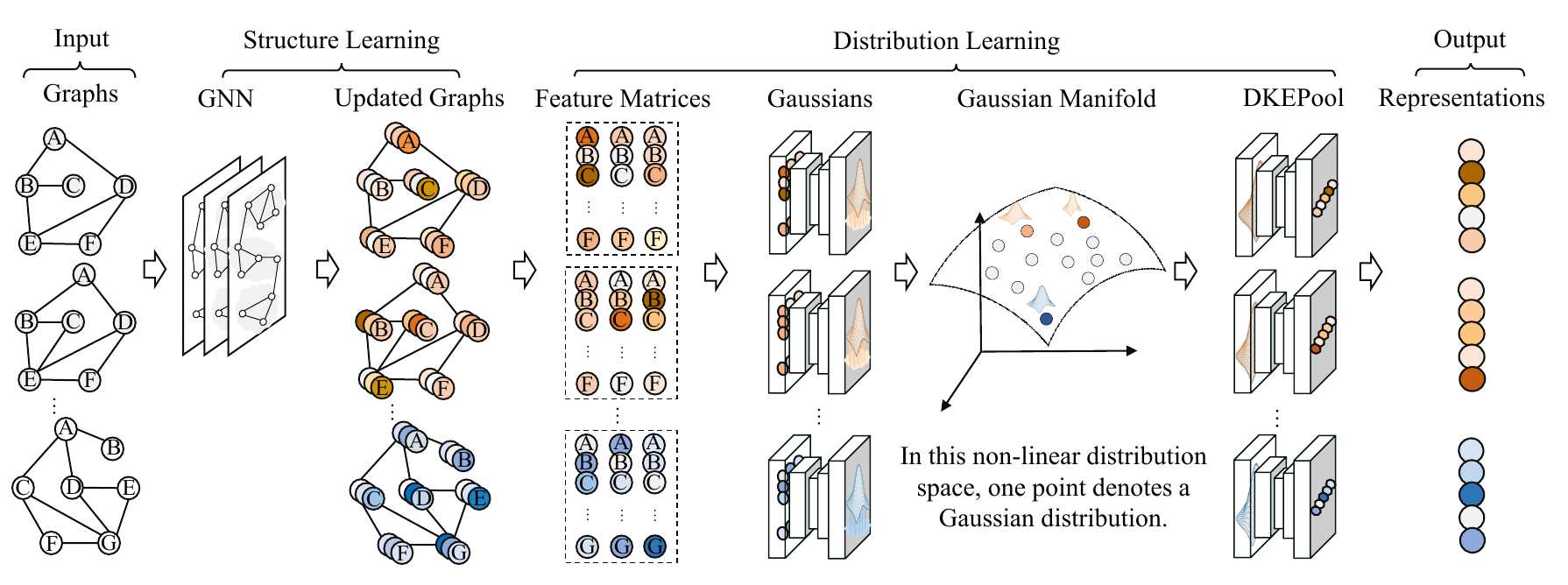}
\caption{A conceptual illustration of the proposed DKEPool network. The graph is viewed as a distribution from which nodes are sampled. DKEPool strives for a vector summarizing the entire distribution that is deemed more informative for downstream graph-level tasks.
}
\label{fig:g2dnet} 
\end{figure*}

A DKEPool network \textit{de facto} disassembles representation learning into two stages, including \textit{structure learning} and \textit{distribution learning}, as shown in Figure~\ref{fig:g2dnet}. Structure learning, powered by existing GNNs, follows a recursive neighborhood aggregation scheme to update node features where graph structure information is absorbed. Distribution learning, on the other hand, omits node interconnections and focuses more on the distribution depicted by all the nodes. Our pooling goal is to learn a vector outlining the node distribution information of the holistic graph, which is used for sake of the following task of graph predictions. Specially, we formulate the graph as a multivariate Gaussian distribution, and thus the distribution space is a special type of Riemannian manifold~\cite{chen2020covariance,chen2019more} that is informative in outlining the non-Euclidean geometry information of graphs.

The contributions of this work are summarized in the following three folds. 

\begin{itemize}[leftmargin = 24 pt]
    \item We argue graph representation learning should consider both graph topology and node distribution, and view graphs as distributions for more efficient graph pooling.
    \item We propose DKEPool, a plug-and-play pooling module that is friendly to existing GNNs to learn informative graph representations.
    \item We provide theoretical analysis to support why our DKEPool can outline distribution information and introduce a robust DKEPool variant that further boosts performance.
\end{itemize}

The remainder of this work is organized as follows. Section~{\ref{Related work}} gives an overview covering the relevant works to this work, including graph pooling methods and Gaussian-based representation methods. In Section~{\ref{Preliminary}}, we present a brief introduction to graph neural networks and two requirements of graph pooling. In Section~{\ref{DKEPool}}, we provide a detailed description of the proposed method. Section~{\ref{Experiments}} provides experimental results on various benchmark datasets to demonstrate the superior performance of the proposed method. Finally, we give a conclusion of this work in Section~{\ref{Conclusion}}.

\section{Related work}\label{Related work}

\textbf{Graph pooling methods.}\; 
In recent years, different approaches to graph representation learning via pooling modules have been investigated in the graph classification tasks. They can be divided into two schools: hierarchical graph pooling and flat graph pooling. 
\emph{Hierarchical graph pooling} is similar to the downsampling strategy in computer vision to gradually coarsen graphs. The works in~\cite{ying2018hierarchical,BianchiGA20,ning2021HAP,yang2020hierarchical} are the most representative hierarchical graph pooling methods. 
DIFFPool~\cite{ying2018hierarchical} proposed a differentiable graph pooling module that can generate hierarchical representations of graphs and can be combined with various graph neural network architectures in an end-to-end fashion. 
MinCutPool~\cite{BianchiGA20} formulated a continuous relaxation of the normalized \texttt{minCUT} problem to addresses limitations of spectral clustering and proposed a graph pooling operator that overcomes some important limitations of the existing graph pooling techniques. 
HAP~\cite{ning2021HAP} utilized a novel cross-level attention mechanism to focus more on close neighborhood  and learned a global graph content  to extract the graph pattern properties for the global guidance in graph coarsening.
\emph{Flat graph pooling}, also known as graph readout operation~\cite{xu2018powerful} plugged at the end of GNN layers, generates single vectors for representing graphs.  Different from hierarchical graph pooling methods, flat graph pooling can only be used once to learn graph representations. The most familiar averaging and summation operations in GNNs~\cite{xu2018powerful,duvenaud2015convolutional,defferrard2016convolutional} belong to the flat graph pooling but only collect the first-order statistic information of the node representations and ignore the important higher-order statistics. Therefore, Wang \emph{et al.}~\cite{wang2020second} proposed a second-order pooling framework as graph pooling and demonstrated the effectiveness and superiority of second-order statistics for graph neural networks.
However, all these methods ignore the non-Euclidean geometry characteristic of graphs, which will cause severe information missing and downgrade the performance of graph representations.

\textbf{Gaussian-based representation methods.}\; 
In statistics, the Gaussian distribution is a type of continuous probability distribution for a real-valued random variable. By considering two main parameters of multivariate Gaussians, \emph{i.e.}, mean and covariance components, the Gaussian-based representation  methods~\cite{nakayama2010global,matsukawa2019hierarchical,li2016local,wang2017g2denet,chen2020covariance,
calvo1990distance,lovric2000multivariate,wang2016raid,li2017second,li2018towards,wang2020deep} have been shown to offer powerful representations for various recognition and regression tasks. Nakayama \textit{et al.} \cite{nakayama2010global} embedded Gaussians in a flat manifold by taking an affine coordinate system and applied them to scene categorization. Calvo \textit{et al.} \cite{calvo1990distance} embedded the manifold of multivariate Gaussians with informative geometry into the manifold of symmetric positive definite matrices with the Siegel metric.
Lovri$\acute{c}$ \textit{et al.}~\cite{lovric2000multivariate} embedded Gaussians in the Riemannian symmetric space~\cite{matsukawa2019hierarchical} with a slightly different metric on the space of multivariate Gaussians. Chen \textit{et al}.~\cite{chen2020covariance} considered the geometry structure of the Riemannian manifold of Gaussians and generated the final representations via Riemannian local difference vectors for image set classification. Wang \textit{et al.}~\cite{wang2016raid} proposed a robust estimation of approximate infinite-dimensional Gaussian based on von Neumann divergence and applied it to material recognition. 
Li \textit{et al.}~\cite{li2016local} proposed the local Log-Euclidean multivariate Gaussian descriptor by defining a multiplication operation on Riemannian manifold to embed Gaussians in the linear space for image classification. Wang \textit{et al.}~\cite{wang2017g2denet,wang2020deep} proposed a novel trainable layer to obtain the square root form of the Gaussian embedding matrix, which considered the Lie group structure of Gaussians and achieved the competitive performance on the challenging fine-grained recognition tasks.
Different from the above works handling Gaussians with complicated Riemannian operations, our designed pooling module embeds Gaussian in the linear space with Euclidean operations while outlining the distribution information of Gaussians.

\section{Preliminary}\label{Preliminary}
A graph is a data structure consisting of two components, \emph{i.e.}, \emph{vertices}, and \emph{edges}. Formally, a graph consisting of $n$ nodes can be represented as $G=(\bm{A},\bm{X})$, where $\bm{A} \in \mathbb{R}^{n\times n}$ is adjacency matrix and $\bm{X}\in \mathbb{R}^{n\times d}$ is the node feature matrix. Given a set of labeled graphs $\mathcal{S} = {(G_1,y_1),(G_2,y_2),...,(G_N,y_N)}$, where $y_i \in \mathcal{Y}$ is the label of $i$-th graph $G_i \in \mathcal{G}$, the purpose of graph-level tasks, \textit{e.g.,} graph classification, is to learn a mapping function $f:\mathcal{G}\rightarrow \mathcal{Y}$ that maps graphs to the set of labels. Different from nodel-level tasks, which mainly leverage the
graph neural networks to generate node representations for downstream tasks, graph-level tasks require holistic graph-level representations for graph-structured inputs of which the size and topology are varying.
To efficiently tackle graph classification tasks, graph neural networks and graph pooling methods are developed to learn powerful node features and graph-level representations.


Graph neural networks (GNNs) are deep learning based models and have recently become widely applied graph analysis methods. The structure learning of the graph is reflected in the aggregation strategy by considering adjacency information.
Given a graph consisting of $n$ nodes $G=(\bm{A},\bm{X})$, GNNs generally follow a message-passing architecture:
\begin{equation}
\label{massagepassing}
\begin{split}
 \bm{H}^{(k)}=M(\bm{A},\bm{H}^{(k-1)};{\theta}^{(k)}),
\end{split}
\end{equation} 
where $\bm{H}^{(k)}$ is the node features of the $k$-th layer and $M$ is the message propagation function. 
The trainable parameters are denoted by $\theta^{(k)}$ and the adjacency matrix by $\bm{A}$. $\bm{H}^{(0)}$ is initialized as $\bm{H}^{(0)}=\bm{X}$. 
The propagation function $M$ can be implemented in various manners~\cite{cai2020multi,kipf2016semi,xu2018powerful,yang2020factorizable}. 
The recursive neighborhood aggregation scheme enables the node embeddings to absorb the structure information. The graph pooling is to obtain the graph-level  representation vector $\bm{h_G}$ for node embeddings $\bm{H}$:
\begin{equation}
\label{pooling}
\begin{split}
 \bm{h_G}=g([\bm{A}],\bm{H}),
\end{split}
\end{equation} 
where $g(\cdot)$ denotes the graph pooling function and $[\bm{A}]$ is the information from $\bm{A}$ can be used in graph pooling. 
Note that $g(\cdot)$ need to satisfy \textbf{two requirements}. \emph{Firstly,} $g(\cdot)$ should be able to take $\bm{H}$ with variable number of rows as the inputs and produce the fixed-sized outputs. \emph{Secondly,} $g(\cdot)$ should output the same represeatation when the the order of nodes of an input graph changes, \emph{i.e.}, the order of rows in $\bm{H}$.


\section{Distribution knowledge embedding for graph-level representation}\label{DKEPool}

The topology information has been learned into node features via GNNs, we argue that the distribution knowledge is also informative for downstream tasks. We first formulate the distributions underlying node features as a multivariate Gaussian distribution, then delineate the proposed distribution learning framework for graph-level representation learning. Finally, we introduce a robust variant based on robust covariance estimation. 

\subsection{Distribution model and representation}

Formally, we can achieve the node embedding features learned from a GNN: $\bm{H} = [\bm{h}_1,\bm{h}_2,...,\bm{h}_n]^T \in \mathbb{R}^{n  \times f}$, where rows of $\bm{H}$, $\bm{h}_i\in\mathbb{R}^{f}, i=1,..,n$, are representations
of $n$ nodes and $f$ denotes the feature dimension that depends on the architecture of GNNs. In this work, we assume the node features follows a multivariate Gaussian distribution:
\begin{equation}
\label{gaussianmodel}
\begin{split}
\mathop P(\bm{h}_i) \sim  \mathcal{N}(\bm{h}_i| \bm{\mu}, \bm{\Sigma}).
\end{split}
\end{equation} 
By the traditional maximum likelihood estimation (MLE) method, the estimation of $\bm{\mu}\in \mathbb{R}^{f}$ and $\bm{\Sigma}\in \mathbb{R}^{f\times f}$ are the mean vector and covariance matrix of samples respectively. These two estimated components are vital to be utilized for Gaussian-based representation learning. 

The representation of prevailing Gaussian-based methods mainly follows two forms,~\textit{vector concatenating}~\cite{nakayama2010global,dai2017fason,chen2020covariance} and \textit{symmetric matrix space embedding}~\cite{matsukawa2019hierarchical,li2016local,wang2020deep,wang2016raid,wang2017g2denet,calvo1990distance,lovric2000multivariate}. In vector concatenating methods, the final representation follows the form of  $[\texttt{vec}(\bm{\Sigma+\mu \mu^T}),\bm{\mu}^T]^T \in \mathbb{R}^{f(f+1)}$~\cite{nakayama2010global}, where \texttt{vec} denotes matrix vectorization. On the other hand, symmetric matrix space embedding methods embedded Gaussian into a symmetric matrix space using a mapping function $\psi$~\cite{calvo1990distance}:
\begin{equation}
\label{Gaussianembedding}
\begin{split}
\mathop \psi(\mathcal{N}(\bm{\mu},\bm{\Sigma}))= \left[  \begin{array}{cc}
      \bm{\Sigma} +\bm{{\mu}{\mu}}^T & \bm{\mu} \\
      \bm{{\mu}}^T & 1
    \end{array} \right],
\end{split}
\end{equation} 
where the resulting representation $\psi(\mathcal{N}(\bm{\mu},\bm{\Sigma}))\in \mathbb{R}^{(f+1)\times(f+1)}$ is a symmetric matrix by operating Siegel metric on the Gaussian manifold. 
Although the above Gaussian-based representation methods have been shown to offer powerful representations for various tasks, they commonly studied Gaussian with complicated Riemannian operations like matrix exponential and logarithm operation, which play a crucial role in the matrix group theory but amidst substantial computational complexity. In contrast to these methods, we introduce an embedding framework with Euclidean operations to outline the distribution information of Gaussians.

\subsection{Distribution knowledge embedding}

In this subsection, we introduce our proposed \textit{Distribution Knowledge Embedding}~(DKEPool) to embed the Gaussians into the linear space with the distribution information preserved. Furthermore, we provide theoretical analysis and proof to support why our DKEPool can outline distribution information and serve as graph pooling.\\

\textbf{DKEPool.}\;
The proposed DKEPool method is inspired by the tremendous success of the bilinear model \cite{yu2018beyond}, which uses several projection matrices to map two vectors into a vector space.  Different from the two input vectors of same size in bilinear model, we need to design a mapping function to address the mean vector $\bm{\mu} \in \mathbb{R}^{f}$ and covariance matrix $\bm{\Sigma} \in \mathbb{R}^{f \times f}$ obtained from node embedding matrix $\bm{H}$. In order to effectively solve the problem of inputs with different size, we define the DKEPool as: $z_i = \texttt{tr}(\bm{\mu} \bm{w}_i^T \bm{\Sigma})$, where $\texttt{tr}$ indicates trace operation on symmetric matrix, $\bm{w}_i \in \mathbb{R}^{f}$ is the $i$-th projection matrix needed to be learned, and $z_i \in \mathbb{R}$ is the output of our proposed module. According to the definition of trace operation on symmetric matrix, the proposed DKEPool can be represented as the following form:
\begin{equation}
\label{}
\begin{split}
z_i = \bm{\texttt{tr}}(\bm{\mu}\bm{w}_i^T \bm{\Sigma}) = \bm{\texttt{tr}}( \bm{w}_i^T \bm{\Sigma \mu}) = \bm{w}_i^T \bm{\Sigma \mu}.
\end{split}
\end{equation}
In order to further obtain the $d$-dimensional representation $\bm{z} = [z_1,z_2,...,z_d] \in \mathbb{R}^{d}$ for describing the entire graph, we need to learn a trainable matrix $\bm{W} = [\bm{w}_1,\bm{w}_2,...,\bm{w}_d] \in \mathbb{R}^{f \times d}$ for different dimensions. To this end, the final $d$-dimensional representation can be obtained by:
\begin{equation}
\label{finalvector}
\begin{split}
\bm{z} = \bm{W}^T \bm{\Sigma \mu},
\end{split}
\end{equation}
where the resulting graph representation form is the covariance matrix multiplied by the mean vector and then projected by a trainable matrix $W$. The DKEPool can outline the distribution information of the Gaussians because it is the generalized form of the \textit{covariance mapping}  mean vector, as presented in the following proposition 1.\\

\begin{figure}
\vspace{-0.5cm}  
\hspace{-10mm}
  \centering
  
  \includegraphics[width=0.86\linewidth]{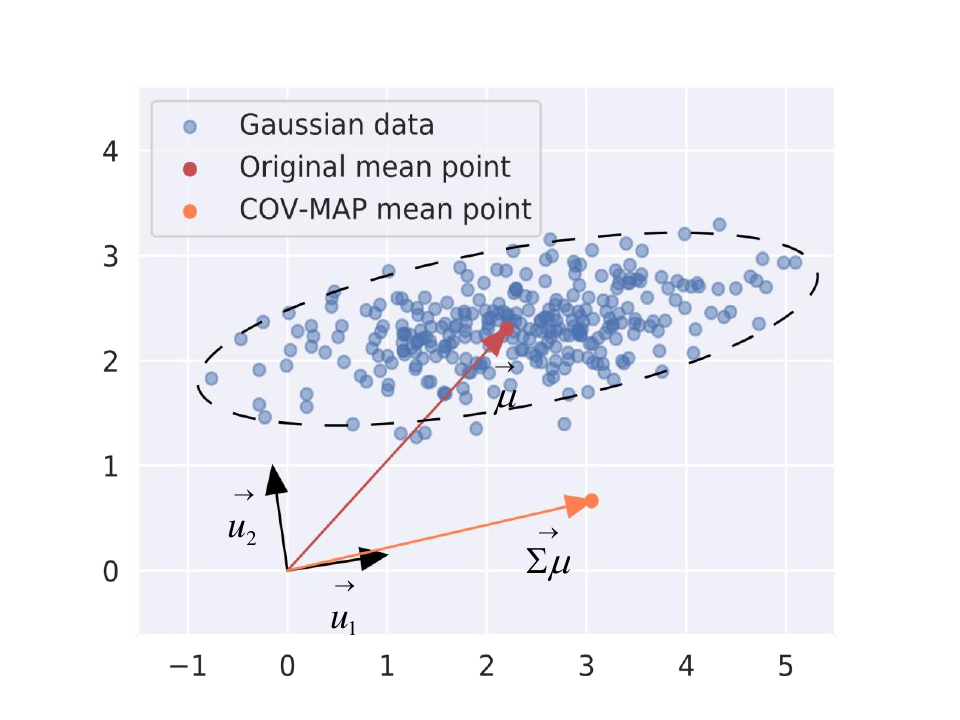} 
  \caption{\label{fig_vector}
	An  illustration of mean vector $\vec{\bm{\mu}}$ and COV-MAP mean vector  $\vec{\bm{\Sigma} \bm{\mu}}$ in 2-dimensional Euclidean space. The $\vec{ \bm{u_1}}$ is the eigen-vector corresponding to bigger eigen-value of covariance matrix and $\vec{\bm{u}}_2$ is another eigen-vector of covariance matrix.}
\end{figure}

\noindent \textbf{Proposition 1.}\; 
\emph{In the Gaussian setting}, the \emph{covariance mapping} (COV-MAP) mean vector $\varphi(\mathcal{N}(\bm{\mu},\bm{\Sigma}){)} = \bm{\Sigma\mu}\in \mathbb{R}^{f}$ can outline the distribution information of Gaussians.  

\noindent \textbf{\textit{Analysis.}}\; From the perspective of mean vector reconstruction in the space spanned by the eigen-vectors of the covariance matrix, the covariance-mapping (COV-MAP) mean vector can be rewritten as:
\begin{equation}
\begin{split}
\bm{\Sigma \mu} =  \bm{U} \bm{\Lambda} \bm{U}^T  \bm{\mu},
\end{split}
\end{equation}
where the diagonal matrix $\bm{\Lambda}={\rm diag}(\lambda_1,\lambda_2,...,\lambda_f)$ consists of the ordered eigen-values of covariance matrix, and orthogonal matrix $\bm{U}=[\bm{u}_1,\bm{u}_2,...,\bm{u}_f]$ consists of the normalized eigen-vectors corresponding their eigen-values. In a new linear space spanned by $\bm{U}$, the projected mean vector can be represented as $\hat{\bm{\mu}} = {\sum}_{i=1}^{f} {\alpha}_i \bm{u}_i = \bm{U} \bm{\alpha}$, where $\bm{\alpha} = [\alpha_1,...,\alpha_f]^T \in \mathbb{R}^{f}$ is the coefficient vector related to the basis vectors $[\bm{u}_1,\bm{u}_2,...,\bm{u}_f]$. 
According to the orthogonal decomposition theorem\footnote{\url{https://mathworld.wolfram.com/OrthogonalDecomposition.html}}, the vector $\bm{u}$ in $\mathbb{R}^{f}$ can be written uniquely in the form of $\bm{\mu} = \bm{\hat{\mu}} + \Delta \bm{\mu}$. 
Note that, as $\Delta \bm{\mu}$ is perpendicular to the space spanned by $\bm{U}$,  $\Delta \bm{\mu}^T \bm{\Sigma}$ will always be zero~\cite{li2019distribution}. Then, COV-MAP mean vector can be represented as: 
\begin{equation}
\begin{split}
\bm{\Sigma \mu} =  \bm{U} \bm{\Lambda} \bm{U}^T  \bm{U} \bm{\alpha} =  \sum_{i=1}^{f}\lambda_i \alpha_i \bm{u}_i,
\end{split}
\end{equation}
where the resulting COV-MAP mean vector is the weighted version of $\bm{\hat{\mu}}$, and weights are eigen-values. 
Considering the largest eigenvalue $\lambda_1$, its corresponding eigen-vector $\bm{u_1}$ reflects the direction of maximum variance and represents the main distribution direction of data. Thus, the COV-MAP mean vector can capture the principal component of data distribution information by using eigen-values as weights. Figure~\ref{fig_vector} shows that the original mean vector of Gaussian data  with purple arrow ($\vec{\bm{\mu}}$) and COV-MAP mean vector with maroon orange ($\vec{\bm{\Sigma \mu}}$). As shown in Figure~\ref{fig_vector}, the COV-MAP mean vector not only contains the information of the mean component but also indicates the distribution information of Gaussians, which is a significant support to our methods. 

Furthermore, we give~\emph{the theoretical proof to check that the resulting vector $\bm{\Sigma\mu}$ can satisfy the two requirements} mentioned in Section~\ref{Preliminary}. For the node embedding features $\bm{H} \in \mathbb{R}^{n  \times f}$ of a graph with $n$ nodes, the mean vector $\bm{\mu}=\frac{1}{n}{\Sigma}_{k=1}^n \bm{h}_k \in \mathbb{R}^{f}$ obviously meets the two requirements. The covariance matrix $\bm{\Sigma} = \frac{1}{n}{\sum}_{k=1}^{n} {(\bm{h}_k-\bm{\mu})}^{T}(\bm{h}_k-\bm{\mu}) \in \mathbb{R}^{f \times f} = \tilde{\bm{H}}^T \tilde{\bm{H}}$, where $\tilde{\bm{H}}$ is the mean centered feature matrix, can satisfy the two requirements as presented in the following proposition 2 and proposition 3. \\

\noindent \textbf{Proposition 2.}\; The covariance operation always outputs an $f\times f$ matrix for $\bm{H} \in \mathbb{R}^{n\times f}$, regardless of the value of $n$.

\noindent \textbf{\textit{Proof.}} The result is obvious since the dimension of $\tilde{\bm{H}}^T\tilde{\bm{H}}$ does not depend on $n$.\\

\noindent \textbf{Proposition 3.}\; The covariance matrix is invariant to permutation so that it outputs the same matrix when the order of rows of feature matrix changes.

\noindent \textbf{\textit{Proof.}} Consider $\bm{H}_1=\bm{P}\bm{H}_2$, where $\bm{P}$ is a permutation matrix. Note that we have $\bm{P}^T\bm{P}=\bm{I}$ for any permutation matrix. Therefore, it is easy to derive: 

\begin{equation}
\begin{split}
{\rm COV}(\bm{H}_1) &=  \tilde{\bm{H}}_1^T\tilde{\bm{H}}_1 = {(\bm{P}\tilde{\bm{H}}_2)}^T{\bm{P}\tilde{\bm{H}}_2}  \\
&=\tilde{\bm{H}}_2^T\bm{P}^T\bm{P}\tilde{\bm{H}}_2\\
&=\tilde{\bm{H}}_2^T\tilde{\bm{H}}_2={\rm COV}(\bm{H}_2).
\end{split}
\end{equation}

As proofed above, the mean and covariance representations satisfy the two requirements to serve as graph  pooling. Therefore, our proposed DKEPool naturally satisfies the two requirements. In order to enhance the performance of DKEPool, we will introduce its robust variant in the following sub-section.


\subsection{Robust distribution knowledge embedding}

Multivariate Gaussian distribution for describing samples has been applied to a large number of recognition tasks. However, it will not work well when the samples are relatively sparse, which means that it is not robust to high dimensional features with a small number of samples. The node samples are sparse in most of graph-level tasks, we thus introduce the robust variant of DKEPool to tackle this problem by considering the robust estimation of the covariance component~\cite{wang2016raid,li2017second,li2018towards,wang2020deep} in the high dimensional feature space.

Wang \textit{et al.}~\cite{wang2016raid} proposed a regularized maximum likelihood estimate (MLE) method based on von Neumann divergence \cite{dhillon2008matrix} for the robust estimation of covariance matrix~(RE-COV). Then, Li \textit{et al.}~\cite{li2017second} proposed a matrix power normalized covariance~(MPN-COV) method to amount to RE-COV for end-to-end training. Due to the limited support of eigen-value decomposition and singular value decomposition on GPU, Li \textit{et al.}~\cite{li2018towards} proposed a fast end-to-end training method based on Newton-Schulz iteration, called iterative matrix square root~(iSQRT) normalization of covariance. 
For the covariance matrix $\bm{\Sigma}$, its robust estimation is represented as $\hat{\bm{\Sigma}}$. Initializing $\bm{Y}_0=\bm{\Sigma}$ and $\bm{Z}_0=\bm{I}$, and the Newton-Schulz iteration in iSQRT takes the following form:
\begin{equation}
\label{iSQRT}
\begin{split}
\bm{Y}_k &= \frac{1}{2}\bm{Y}_{k-1}(3\bm{I}-\bm{Z}_{k-1}\bm{Y_}{k-1}) \\
\bm{Z}_k &= \frac{1}{2}(3\bm{I}-\bm{Z}_{k-1}\bm{Y}_{k-1})\bm{Z}_{k-1},
\end{split}
\end{equation}
where $\bm{Y}_k$ is the resulting robust estimation of the covariance component after $k$ iterations, \emph{i.e.} $\hat{\bm{\Sigma}}=\bm{Y}_k$. Note that the derivation of backward propagation in iSQRT\footnote{Interested readers please refer to~\cite{wang2020deep,li2018towards} for details.} is not straightforward, even though autograd toolkits provided by some deep learning frameworks can accomplish this task automatically, \emph{e.g.}, autograd of PyTorch. The partial derivatives of the loss function $l$ with respect to $\bm{Y}_k$ and $\bm{Z}_k$ are as follows:
\begin{equation}
\label{iSQRT_BP}
\begin{split}
\frac{\partial l}{\partial \bm{Y}_{k-1}} &=\frac{1}{2}(\frac{\partial l}{\partial \bm{Y}_{k}}(3\bm{I}-\bm{Y}_{k-1}\bm{Z}_{k-1}) - \bm{Z}_{k-1}\frac{\partial l}{\partial \bm{Z}_{k}}\bm{Z}_{k-1}  			\\ 
&  -  \bm{Z}_{k-1}\bm{Y}_{k-1}\frac{\partial l}{\partial \bm{Y}_{k}}) 		\\
\frac{\partial l}{\partial \bm{Z}_{k-1}} &=\frac{1}{2}((3\bm{I}-\bm{Y}_{k-1}\bm{Z}_{k-1})\frac{\partial l}{\partial \bm{Z}_{k}} - \bm{Y}_{k-1}\frac{\partial l}{\partial \bm{Y}_{k}}\bm{Y}_{k-1}  			\\ 
&  -  \frac{\partial l}{\partial \bm{Z}_{k}}\bm{Z}_{k-1}\bm{Y}_{k-1}).
\end{split}
\end{equation}

In this work, we adopt the iSQRT to improve our proposed framework due to its friendly performance on GPU~\cite{li2018towards} and its original version considered in the Gaussian setting~\cite{wang2016raid}. The robust DKEPool can be represented as:
\begin{equation}
\label{robust_rep}
\begin{split}
\bm{z} = \bm{W}^T \hat{\bm{\Sigma}} \bm{\mu},
\end{split}
\end{equation}
the above equation Eq.(\ref{robust_rep}) is the robust variant of our proposed DKEPool resulting from iSQRT. The difference between Eq.(\ref{finalvector}) and Eq.(\ref{robust_rep}) only exists in the covariance component.

\section{Experiments}\label{Experiments}

In this section, we demonstrate the experimental results of our proposed DKEPool module and its robust variant DKEPool$_R$ on graph classification datasets. To verify the effectiveness of our methods, we first perform comparison experiments with existing benchmark graph classification methods, including kernel-based and GNN-based methods, and then compare our modules with the Gaussian-based representation methods by using the same GNN. Secondly, we apply the DKEPool module to hierarchical methods for stepwise information extracting. Finally, we provide the ablation studies in the section \ref{ablation_study}.

\textbf{Datasets.} We adopt seven popularly used \textbf{TU databases}~\cite{morris2020tudataset} in this paper, including IMDB-BINARY~(IMDB-B), IMDB-MULTI~(IMDB-M), MUTAG,  PTC, NCI1, PROTEINS and REDDIT-BINARY~(RDT-B), and two more recently released Open Graph Benchmark~(\textbf{OGB}) \textbf{datasets}~\cite{hu2020open}, including OGB-MOLHIV and OGB-MOLBBBP. The MUTAG, PTC, PROTEINS, NCI1, OGB-MOLHIV and OGB-HIVBBBP are bioinformatics datasets where each graph represents a chemical compound. On the other side, the IMDB-B, IMDB-M, RDT-B are social datasets. Below are the detailed descriptions of datasets:

\begin{itemize}[leftmargin = 24 pt]
\item IMDB-B is a movie collaboration dataset of 1,000 graphs representing ego-networks for actors/actresses. The dataset is derived from collaboration graphs on action and romance genres. In each graph, nodes represent actors/actresses and edges simply mean they collaborate on the same movie.
\item IMDB-M is the multi-class version of IMDB-BINARY. It contains 1,500 ego-networks and has three extra genres, namely, Comedy, Romance and Sci-Fi. The
graphs are labeled by the corresponding genre and the task is to identify the genre for each graph.
\item MUTAG is a mutagenic aromatic and heteroaromatic nitro compounds dataset and their graph label indicates whether the mutagenic effect on bacteria exists. It contains 188 graphs and seven discrete node labels. The task is to classify whether the compound is aromatic or heteroaromatic.
\item PTC is a chemical compounds dataset that represents the carcinogenicity for male and female rats. It consists of 344 graphs
representing chemical compounds. Each node comes with one of 19 discrete node labels. The task is to predict the rodent carcinogenicity for each graph.
\item PROTEINS represents protein structures which are helix, sheet and turn. There are 1,113 graph structures of proteins in this dataset. Nodes in the graphs refer to secondary structure elements (SSEs) and edges mean that two nodes are neighbors along the amino-acid sequence or in space.
\item RDT-B is a dataset of 2,000 graphs where each graph represents an online discussion thread. Datasets are crawled from top submissions under four popular subreddits. Nodes denote users in the corresponding discussion thread and an edge means that one user responded to another.
\item NCI1 is a bioinformatics dataset of 4,110 graphs representing chemical compounds used for activity against non-small cell lung cancer cell lines. It contains data published by the National Cancer Institute (NCI). Each node is assigned with one of 37 discrete node labels.
\item OGB-MOLHIV is a property prediction dataset of different sizes that adopted from the MOLECULENET. Input node features, contain atomic number and chirality, as well as other additional atom features. It contains 41,127 graphs with an average of 25.51 nodes in each graph.
\item OGB-MOLBBBP is a small dataset from MOLECULENET for binary classification. It also can be used to molecule-specific methods. Each graph represents a molecule, where nodes are atoms, and edges are chemical bonds.  It contains 2,039 graphs with an average of 24.06 nodes in each graph.
\end{itemize}

\begin{table*}
\vspace{-0.1cm}  
\setlength{\abovecaptionskip}{0.1cm} 
\caption{Comparison results between our proposed methods and existing benchmark graph classification methods on TU datasets. DKEPool$_R$ denotes the robust variant of DKEPool. The best models on each dataset are highlighted with \textbf{boldface.}}
\label{TUResults}
\renewcommand{\arraystretch}{1.0} 
\large
\centering 
\resizebox{1.0\textwidth}{!}{
\begin{tabular}{clccccccc}
\toprule
& 		&IMDB-B 	&IMDB-M 	&MUTAG  	&PTC  	&PROTEINS  	&RDT-B  	&NCI1 \\
\midrule
\multirow{4}{*}{\textbf{\rotatebox{90}{Dataset}}}
& \# Graphs   		&1000		&1500		&188		&344		&1113	 	&2000  	&4110 	\\
& \# Classes   	     	&2		&3		&2		&2		&2		&2  		&2 		\\
& Nodes(max)   		&136		&89		&28		&109		&620		&3783  	&111 		\\
& Nodes(avg.)   	&19.8		&13.0		&18.0		&25.6		&39.1		&429.6  	&29.2 	\\

\midrule
\multirow{5}{*}{\textbf{\rotatebox{90}{Kernel}} }
&GK~\cite{shervashidze2009efficient}	 	
			&65.9$\pm$1.0 	&43.9$\pm$0.4 	&81.4$\pm$1.7	&57.3$\pm$1.4	&71.7$\pm$0.6 	&77.3$\pm$0.2	&62.3$\pm$0.3		\\
&RW~\cite{vishwanathan2010graph}
		 	&-			&- 			&79.2$\pm$2.1	&57.9$\pm$1.3	&74.2$\pm$0.4   &-			&-				\\
&WL~\cite{shervashidze2011weisfeiler}
		 	&73.8$\pm$3.9	&50.9$\pm$3.8	&82.1$\pm$0.4	&60.0$\pm$0.5	&74.7$\pm$0.5	&-	        	&82.2$\pm$0.2			\\
&DGK~\cite{yanardag2015deep}
			&67.0$\pm$0.6	&44.6$\pm$0.5	&-			&60.1$\pm$2.6	&75.7$\pm$0.5 	&78.0$\pm$0.4	&80.3$\pm$0.5 		\\
&AWE~\cite{ivanov2018anonymous}
			&74.5$\pm$5.9	&51.5$\pm$3.6	&87.9$\pm$9.8	&-			&- 			&87.9$\pm$2.5	&-				\\

\midrule
\multirow{8}{*}{\textbf{\rotatebox{90}{GNN}} }

&ASAP~\cite{ranjan2020asap}
			&77.6$\pm$2.1	&54.5$\pm$2.1	&91.6$\pm$5.3	&72.4$\pm$7.5 	&78.3$\pm$4.0	&93.1$\pm$2.1	&75.1$\pm$1.5 		\\


&SOPool~\cite{wang2020second}
		 	&78.5$\pm$2.8	&54.6$\pm$3.6 	&95.3$\pm$4.4	&75.0$\pm$4.3	&80.1$\pm$2.7	&91.7$\pm$2.7	&84.5$\pm$1.3		\\

&GMT~\cite{BaekKH21}
			&79.5$\pm$2.5	&55.0$\pm$2.8	&95.8$\pm$3.2	&74.5$\pm$4.1 	&80.3$\pm$4.3	&93.9$\pm$1.9	&84.1$\pm$2.1		\\


&HAP~\cite{ning2021HAP}
			&{79.1$\pm$2.8}	&{55.3$\pm$2.6}	&{95.2$\pm$2.8}	&{75.2$\pm$3.6}	&{79.9$\pm$4.3} &{92.2$\pm$2.5}	&{81.3$\pm$1.8}		\\
&{PAS~\cite{liangwei2021PAS} }
			&{77.3$\pm$4.1}	&{53.7$\pm$3.1}	&{94.3$\pm$5.5}	&{71.4$\pm$3.9}	&{78.5$\pm$2.5} &{93.7$\pm$1.3}	&{82.8$\pm$2.2}		\\
&{HaarPool~\cite{Wang2020haarpool} }
			&{79.3$\pm$3.4}	&{53.8$\pm$3.0}	&{90.0$\pm$3.6}	&{73.1$\pm$5.0}	&{80.4$\pm$1.8} &{93.6$\pm$1.1}	&{78.6$\pm$0.5}		\\
&{DiffPool~\cite{ying2018hierarchical} }
			&{73.9$\pm$3.6}	&{50.7$\pm$2.9}	&{94.8$\pm$4.8}	&{68.3$\pm$5.9}	&{76.2$\pm$3.1} &{91.8$\pm$2.1}	&{76.6$\pm$1.3}		\\
&{GMN~\cite{khasahmadi2020GMN} }
			&{76.6$\pm$4.5}	&{54.2$\pm$2.7}	&{95.7$\pm$4.0}	&{76.3$\pm$4.3}	&{79.5$\pm$3.5} &{93.5$\pm$0.7}	&{82.4$\pm$1.9}		\\

\midrule
\multirow{2}{*}{\textbf{\rotatebox{90}{Ours}}}
&DKEPool		&75.1$\pm$2.5 	&49.6$\pm$3.7	&96.8$\pm$3.5	&77.9$\pm$4.0	&80.5$\pm$4.2 	&\textbf{95.0$\pm$1.0}	&84.7$\pm$1.9 \\	
&DKEPool$_R$	&\textbf{80.9$\pm$2.3} 	&\textbf{56.3$\pm$2.0}	&\textbf{97.3$\pm$3.6}	&\textbf{79.6$\pm$4.0}	&\textbf{81.2$\pm$3.8} 	&94.8$\pm$0.5		&\textbf{85.4$\pm$2.3}\\	

\bottomrule

\end{tabular}
}
\vspace{-0.2cm}
\end{table*}

\begin{table}
\vspace{-0.1cm}  
\setlength{\abovecaptionskip}{0.1cm} 
\caption{Comparison results between our proposed methods and existing baselines on OGB datasets.The best models are highlighted with \textbf{boldface.}}
\label{OGBResults}
\renewcommand{\arraystretch}{1.0} 
\footnotesize
\centering
\resizebox{.48\textwidth}{!}{
\begin{tabular}{lcc}

\toprule
			&MOLHIV  	&MOLBBBP   \\
\midrule
\# Graphs   	&41127		&2039			\\
\# Classes   	&2			&2			\\
Nodes(avg.)   	&25.51		&24.06		\\
\midrule
GCN~\cite{kipf2016semi} 
            	&76.18$\pm$1.26	&65.67$\pm$1.86		\\
GIN~\cite{xu2018powerful}
            	&75.84$\pm$1.35	&66.78$\pm$1.77		\\
\midrule
ASAP~\cite{ranjan2020asap}	
            	&72.86$\pm$1.40	&63.50$\pm$2.47		\\

SOPool~\cite{wang2020second}		
            	&76.98$\pm$1.11	&65.82$\pm$1.66		\\
GMT~\cite{BaekKH21}	
            	&77.56$\pm$1.25	&68.31$\pm$1.62		\\

{HAP~\cite{ning2021HAP} }
			&{75.71$\pm$1.33}		&{66.01$\pm$1.43}		\\
{PAS~\cite{liangwei2021PAS} }
			&{77.68$\pm$1.28}		&{66.97$\pm$1.21}		\\
{HaarPool~\cite{Wang2020haarpool} }
			&{74.69$\pm$1.62}		&{66.11$\pm$0.82}		\\
{DiffPool~\cite{ying2018hierarchical} }
            	&{75.64$\pm$1.86}		&{68.25$\pm$0.96}		\\
{GMN~\cite{khasahmadi2020GMN} }
			&{77.25$\pm$1.70}		&{67.06$\pm$1.05}		\\

\midrule
DKEPool		
			&{77.10$\pm$1.35}	&{67.94$\pm$1.34}	 \\		
DKEPool$_R$
			&\textbf{78.65$\pm$1.19}	&\textbf{69.73$\pm$1.51}	\\	
\bottomrule

\end{tabular}
}
\vspace{-0.65cm}
\end{table}

\begin{table*}
\caption{Comparison results between our proposed methods and other Gaussian-based representation methods by using the same GNN network, The best performers on each dataset are highlighted with \textbf{boldface}.}
\label{gaussian_table}
\renewcommand{\arraystretch}{1.0} 
\LARGE
\centering
\resizebox{\textwidth}{!}{

\begin{tabular}{lccccccc}
\toprule
&		\multicolumn{5}{c}{TU datasets}  & \multicolumn{2}{c}{OGB datasets} \\
\cmidrule(lr){2-6}
\cmidrule(l){7-8}
          &MUTAG  	&PTC  &PROTEINS  &RDT-B  &NCI1	&MOLHIV	&MOLBBBP		 \\
\midrule
GaussING~\cite{nakayama2010global}			
&94.7$\pm$4.1	&74.4$\pm$4.0	&76.3$\pm$4.0	&94.2$\pm$0.8  &84.2$\pm$1.2	&74.2$\pm$1.4	&67.4$\pm$1.3\\
GaussREM~\cite{wang2016raid}			
&95.2$\pm$4.5	&75.5$\pm$6.5	&80.1$\pm$3.4	&94.1$\pm$0.7  &\textbf{85.5$\pm$1.9}	&77.5$\pm$1.3	&67.0$\pm$1.6\\
GaussPRM~\cite{matsukawa2019hierarchical}			
&96.3$\pm$4.7	&75.0$\pm$4.5	&76.8$\pm$3.6	&93.9$\pm$1.1  &84.1$\pm$1.7	&75.0$\pm$1.4	&67.4$\pm$1.4\\
\midrule
GaussVCAT~\cite{dai2017fason}  	
&95.2$\pm$4.5	&75.0$\pm$5.2	&79.2$\pm$3.6	&94.2$\pm$0.9 	&85.3$\pm$1.7	&77.2$\pm$1.5	&66.8$\pm$1.6\\
GaussEMBD~\cite{calvo1990distance}
&94.1$\pm$5.7  	&76.4$\pm$6.5   &77.4$\pm$3.3	&94.1$\pm$0.8	&84.3$\pm$1.6	&75.9$\pm$1.1	&68.0$\pm$1.4\\
DKEPool~(Ours)	
&96.8$\pm$3.5	&77.9$\pm$4.0	&80.5$\pm$4.2 	&\textbf{95.0$\pm$1.0}	&84.7$\pm$1.9	&77.1$\pm$1.4	&67.9$\pm$1.3\\		
\midrule
GaussRVCAT~\cite{chen2020covariance}
&95.7$\pm$4.7	&75.9$\pm$4.9	&80.4$\pm$3.6	&94.8$\pm$0.8  &85.4$\pm$1.9	&76.8$\pm$1.2	&67.7$\pm$1.8\\
GaussREMBD~\cite{wang2020deep}
&96.8$\pm$4.3 	&75.6$\pm$4.2	&79.9$\pm$3.5	&93.9$\pm$0.7  &85.1$\pm$1.7	&76.4$\pm$1.3	&68.3$\pm$1.5\\
DKEPool$_R$~(Ours)		
&\textbf{97.3$\pm$3.6}	&\textbf{79.6$\pm$4.0}	&\textbf{81.2$\pm$3.8}	&94.8$\pm$0.5		&85.4$\pm$2.3	&\textbf{78.7$\pm$1.2}	&\textbf{69.7$\pm$1.5}\\	
\bottomrule
\end{tabular}
}
\end{table*}

\textbf{Settings.} To ensure the non-singularity of the covariance matrix for learning DKEPool$_R$, we add Gaussian noise to node embedding features with a fixed signal-to-noise ratio~(SNR$\in \{10, 15, 20\}$). For the dimension of final representation, we tune $d$ of the projection matrix $W$ in Eq.(\ref{finalvector}) $\in \{200, 400, 600, 800\}$. 
For the number of the iterations in DKEPool$_R$, we follow~\cite{li2018towards} to set  $k$ as 5 on all datasets.
For the \textbf{TU datasets,} the nodes have categorical labels as input features on bioinformatics datasets. 
To be specific, we set all node feature vectors to be the same for RDT-B. For the other social network datasets, we use one-hot encoding of node degrees as features. We follow the experimental settings in~\cite{wang2020second,gao2021topology,gao2019graph} for performance evaluation. To obtain graph node features, we adopt the GIN~\cite{xu2018powerful} and set the number of GIN layer as 5 in our experiments. In particular, we select the batch size $\in$ \{32, 128\}, hidden dimension $\in$ \{16, 32, 64\}, learning rate $\in$ \{$1e-2, 5e-3, 1e-3$\}. 
For the \textbf{OGB datasets,} we follow the experimental settings  in~\cite{BaekKH21} for performance evaluation. To aggregate neighborhood information for updating node features, we use the GCN~\cite{kipf2016semi} framework and set the number of GCN layer as 3 in our experiments. In particular, we select the batch size $\in$ \{128, 512\}, hidden dimension $\in$ \{128, 256\}, learning rate $\in$ \{$1e-3, 5e-4, 1e-4$\} and set the  weight decay to be $1e-4$. 

\subsection{Comparison results with state-of-the-art methods}\label{sotacompare}
For the {comparison experiments}, we compare our method with several state-of-the-art~(SOTA) GNN-based methods, includingPooling~(DiffPool)~\cite{ying2018hierarchical}, Hierarchical Adaptive Pooling~(HAP)~\cite{ning2021HAP}, Second-Order Pooling (SOPool)~\cite{wang2020second}, Pooling Architecture Search~(PAS)~\cite{liangwei2021PAS}, Harr Graph Pooling~(HarrPool)~\cite{Wang2020haarpool}, Differentiable  Graph Memory Network~(GMN)~\cite{khasahmadi2020GMN}, Adaptive Structure Aware Pooling~(ASAP)~\cite{ranjan2020asap} and Graph Multiset Transformer~(GMT)~\cite{BaekKH21} on all datasets. Furthermore, we compare five kernel-based baselines on TU datasets, 
including Graphlet Kernel (GK)~\cite{shervashidze2009efficient}, Random Walk Kernel~(RW)~\cite{vishwanathan2010graph}, Weisfeiler-Lehman subtree kernel~(WL)~\cite{shervashidze2011weisfeiler}, Deep Graphlet Kernel~(DGK)~\cite{yanardag2015deep}, and Anonymous Walk Embeddings~(AWE)~\cite{ivanov2018anonymous}, and two GNN baselines on OGB datasets, including GCN~\cite{kipf2016semi} and GIN~\cite{xu2018powerful}. To ensure the fairness of these comparison experiments, we select the same number of network layers, learning rate, hidden dimension, and batch size as the settings in the DKE framework. Other hyperparameters of different models remain unchanged as the settings in their source codes. 

Table~\ref{TUResults} shows the classification accuracies achieved by our proposed approaches on TU datasets as compared with several aforementioned GNN-based and graph kernel baselines. 
As demonstrated in Table~\ref{TUResults}, our proposed methods achieve the best performances for graph classification on TU datasets. In particular, DKEPool and DKEPool$_R$ achieved better accuracies than others on MUTAG, PTC, PROTEINS, RDT-B and NCI1 datasets. On the other hand, the performance of DKEPool is weaker than some baselines on IMDB-B and IMDB-M datasets.
Table~\ref{OGBResults} provides the recognition results compared with different graph pooling methods on two OGB datasets, \emph{i.e.}, MOLHIV and MOLBBBP. The superior performance demonstrates the effectiveness of our proposed approaches. The ROC-AUC of DEKPool$_R$ on MOLHIV and MOLBBBP improve 1.25 \% and 2.08 \% compared with the second performance in Table~\ref{OGBResults}. Moreover, \emph{p}-values are $4.81e-2$, $2.88e-2$ between DEKPool$_R$ and second performance method on these two datasets, which analyze their performance by considering the mean values and standard deviations and show that the improvement of our approach is significant.


\subsection{Comparison results with other Gaussian-based representations}\label{gaussiancompare}
We have already demonstrated the superiority of our proposed methods over previous baselines. In this sub-section, we compared our framework with other Gaussian-based representation methods, including Gaussian approach using information geometry (GaussING)~\cite{nakayama2010global}, Robust estimation of Gaussian (GaussREM)~\cite{wang2016raid}, Gaussian  parametrized as Riemannian manifold (GaussPRM)~\cite{matsukawa2019hierarchical,lovric2000multivariate}, vector concatenating of mean vector and covariance matrix (GaussVCAT)~\cite{dai2017fason}, vector concatenating of mean vector and robust covariance matrix (GaussRVCAT)~\cite{chen2020covariance}, Gaussian embedding\footnote{In this paper, Gaussian embedding especially  indicates the form of Eq.(\ref{Gaussianembedding}).} (GaussEMBD)~\cite{calvo1990distance}, and robust Gaussian Embedding (GaussREMBD)~\cite{wang2020deep,wang2017g2denet}. 

Table~\ref{gaussian_table} provides the comparison results between different Gaussian-based representation methods by using the same GNN network on OGB-MOLHIV, OGB-MOLBBBP, MUTAG, PTC, NCI, PROTEINS and RDT-B datasets. As shown in Table \ref{gaussian_table}, our proposed methods achieve \emph{competitive} results compared with other Gaussian-based representations. On NCI1 dataset, Gaussian RobustE achieves the best performance, and other Gaussian-based methods also achieve very impressive results.  
Furthermore, the superior performances in Table~\ref{gaussian_table} \emph{vertify the importance of distribution knowledge} in graph node features for graph-level representation learning. 

\begin{table}
\caption{The results of the proposed DKE applied in DiffPool and MinCutPool for stepwise information extracting. The best models are highlighted with \textbf{boldface.}}
\label{table:hierarchical_dke}
\renewcommand{\arraystretch}{1.0} 
\small
\normalsize 
\large
\centering
\resizebox{.5\textwidth}{!}{
\begin{tabular}{lcc}
\toprule
					&PROTEINS	  			&MOLHIV      \\
\midrule
DiffPool  				&76.23$\pm$3.10			&75.64$\pm$1.86\\
DiffPool$_{dke}$  		&78.65$\pm$2.98			&\textbf{77.83$\pm$1.79}\\
DiffPool$_{dkeloss}$   	&\textbf{78.98$\pm$2.72}	&76.71$\pm$1.53\\

\midrule
MinCutPool   			&77.43$\pm$2.64			&75.37$\pm$2.05\\
MinCutPool$_{dke}$   		&78.90$\pm$2.51			&\textbf{77.76$\pm$1.81}\\
MinCutPool$_{dkeloss}$ 	&\textbf{79.30$\pm$2.45}	&77.19$\pm$1.97\\
\bottomrule
\end{tabular}
}
\end{table}


\begin{figure*}
\centering
\hspace{-2mm}
\subfigure[DAS on MUTAG]{\includegraphics[width=4.3cm]{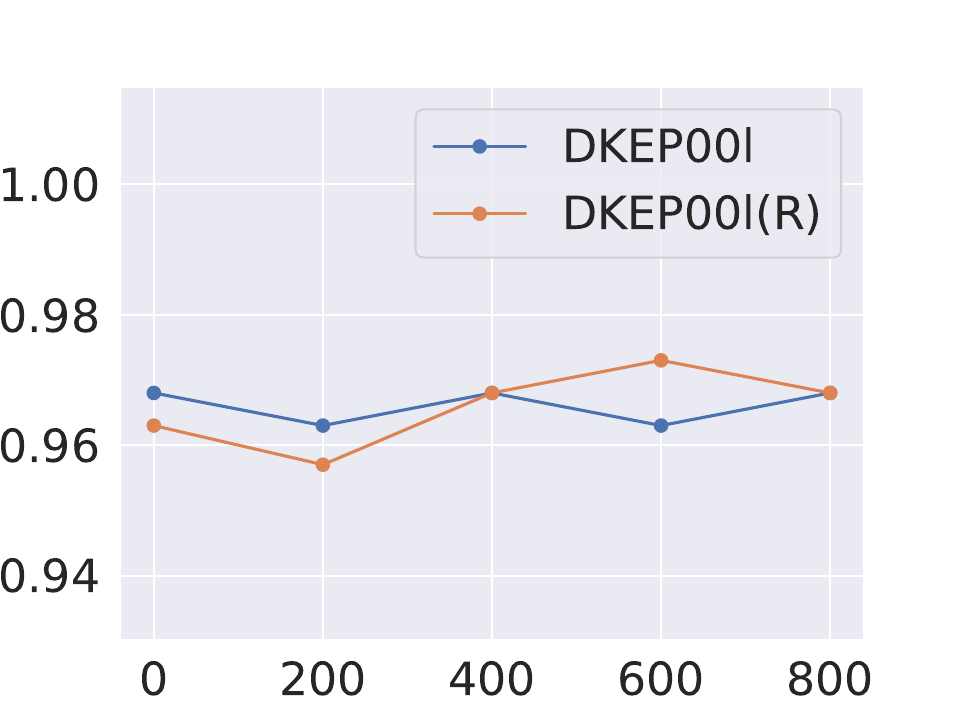}}
\hspace{-0mm}
\subfigure[DAS on PTC]{\includegraphics[width=4.3cm]{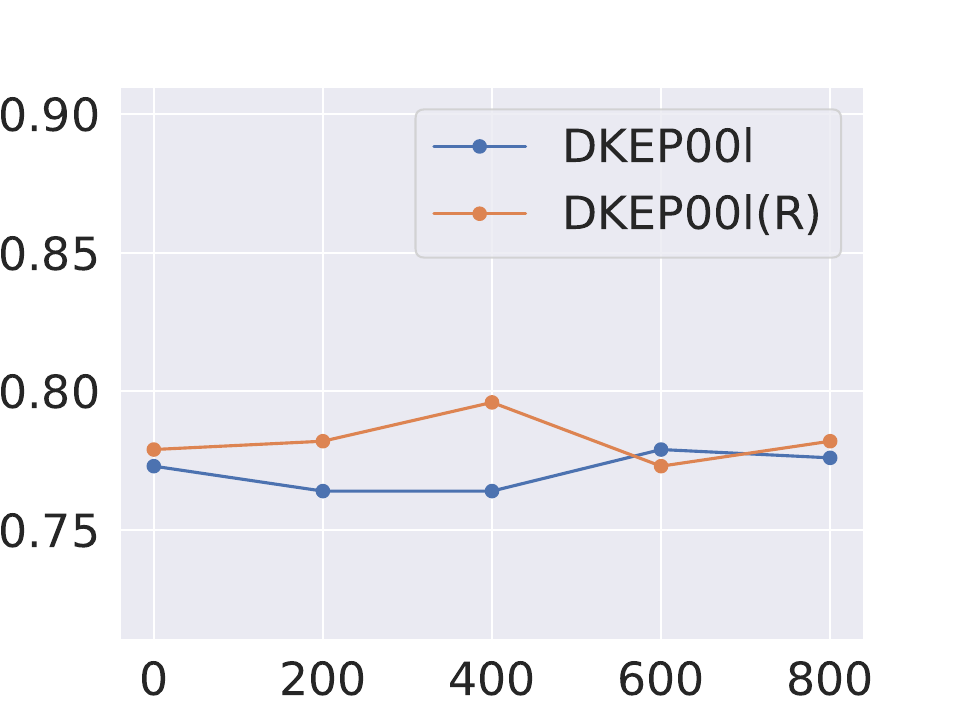}}
\hspace{-0mm}
\subfigure[DAS on PROTEINS]{\includegraphics[width=4.3cm]{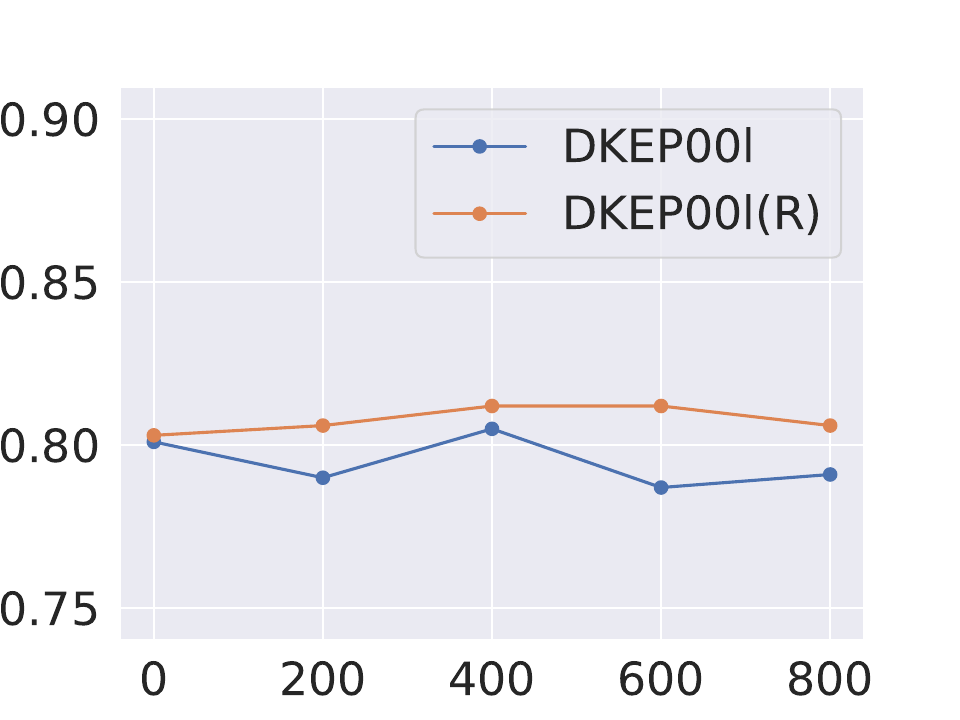}}
\hspace{-0mm}
\subfigure[DAS on RDT-B]{\includegraphics[width=4.3cm]{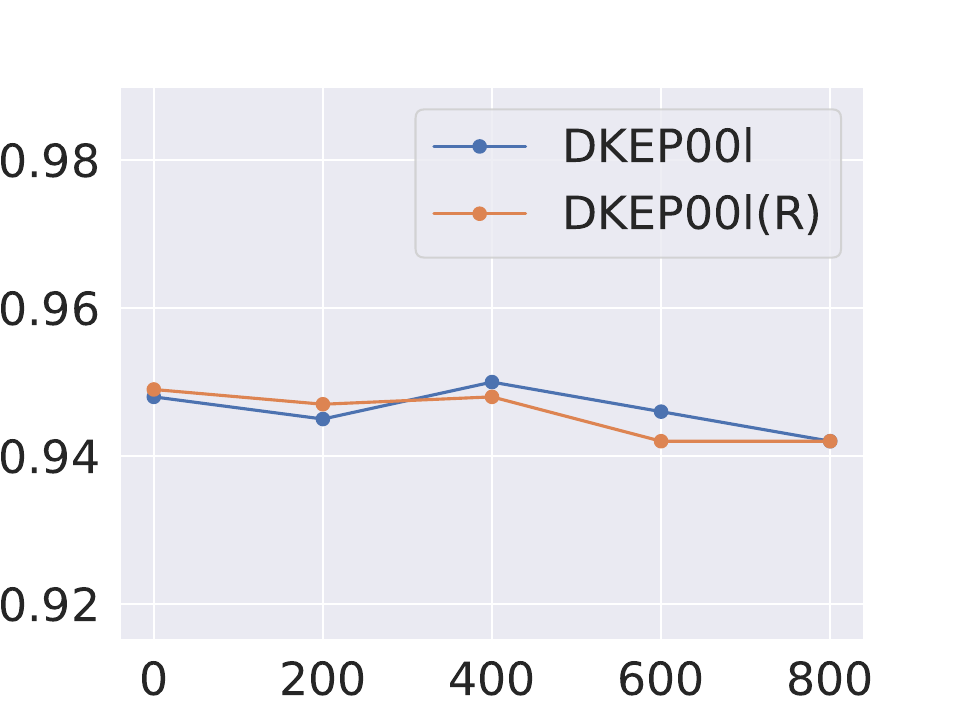}}
\hspace{-2mm}

\vspace{1mm}

\hspace{-2mm}
\subfigure[DAS on NCI1]{\includegraphics[width=4.3cm]{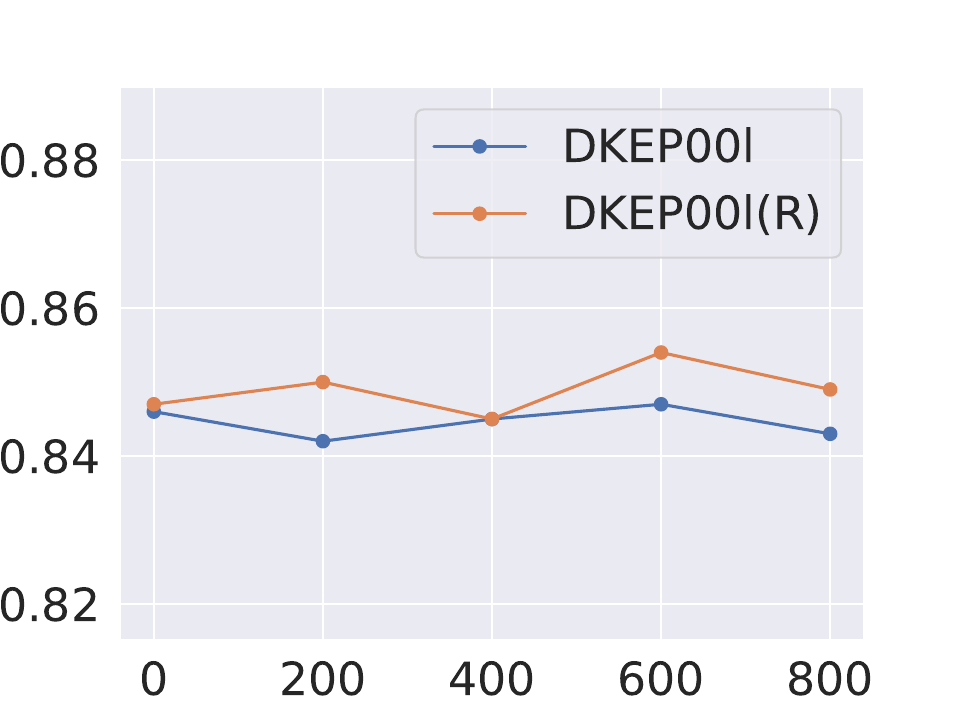}}
\hspace{-0mm}
\subfigure[DAS on IMDB-B]{\includegraphics[width=4.3cm]{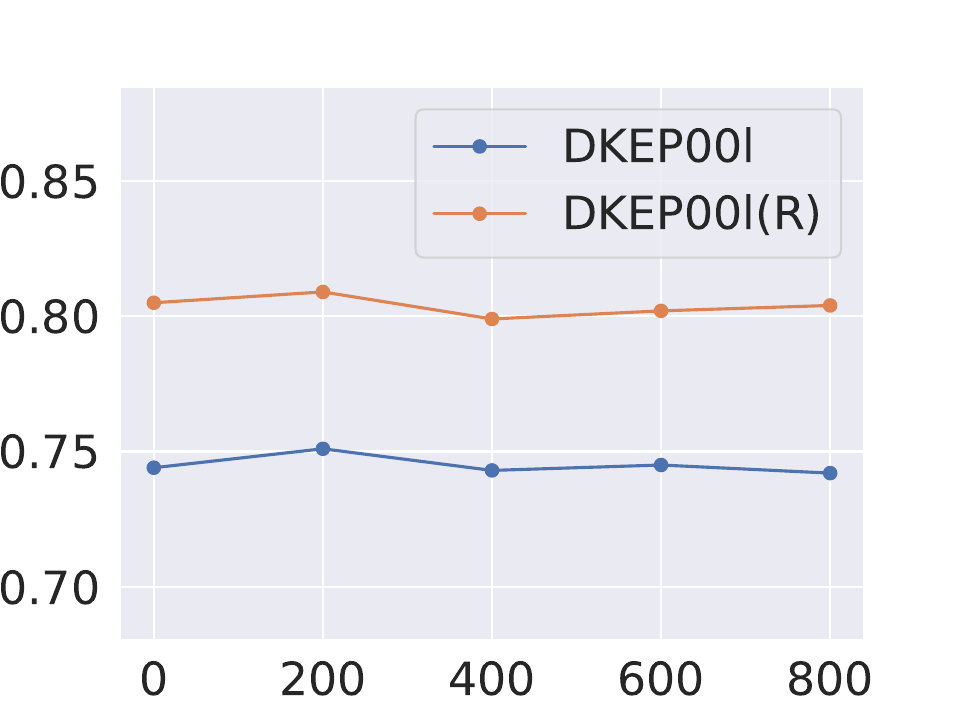}}
\hspace{-0mm}
\subfigure[DAS on IMDB-M]{\includegraphics[width=4.3cm]{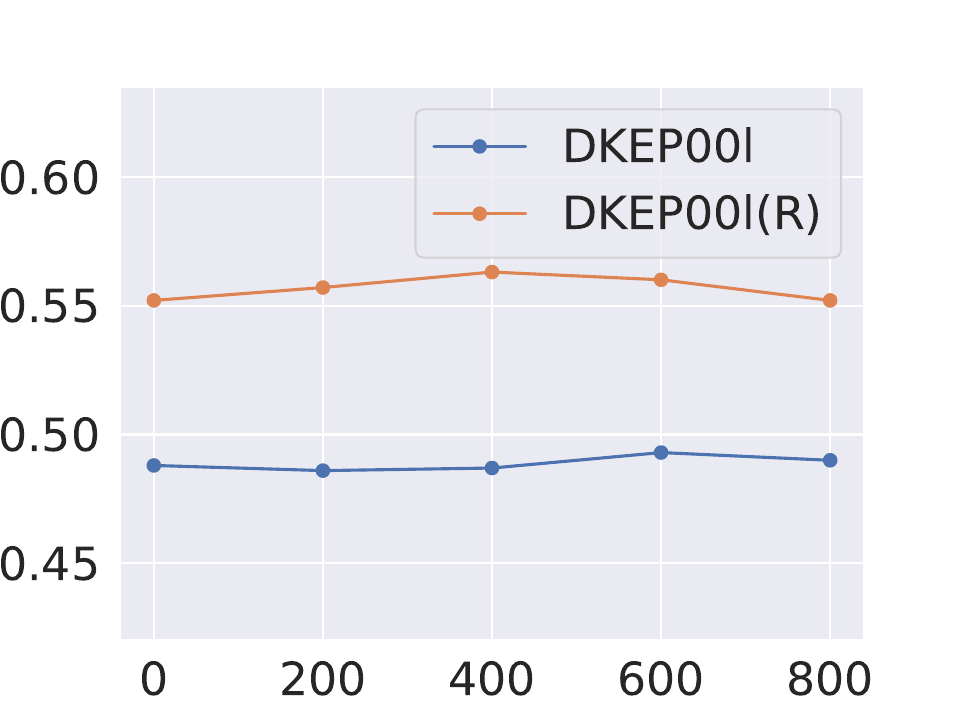}}
\hspace{-0mm}
\subfigure[DAS on MOLBBBP]{\includegraphics[width=4.3cm]{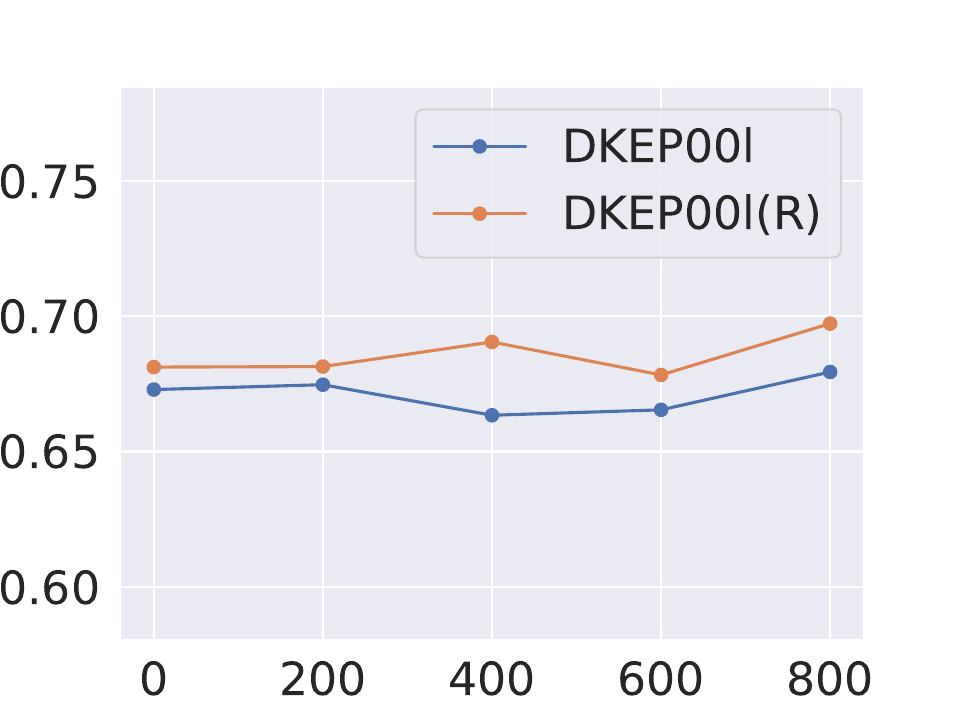}}
\hspace{-2mm}

\caption{\label{das_ablation}
Ablation studies of DKEPool and DKEPool$_R$ in representation dimension.  The abscissa refers to the representation dimension $d$, and ordinate refers to the recognition rates.
}
\vspace{-0.0cm}  
\end{figure*}

\begin{figure*}
\centering
\hspace{-2mm}
\subfigure[SAS on MUTAG]{\includegraphics[width=4.3cm]{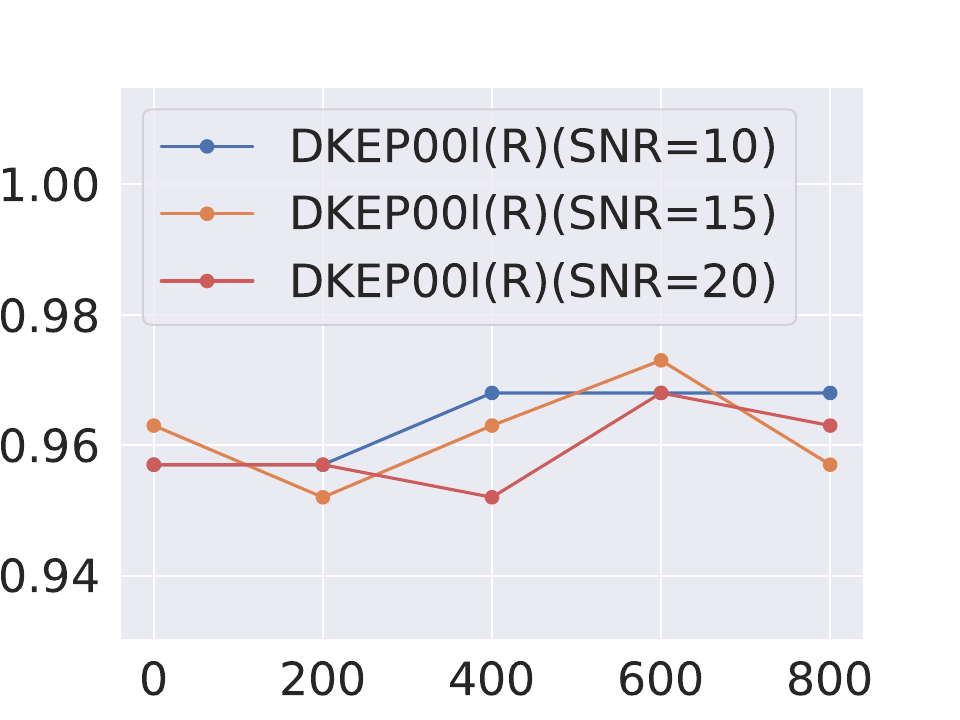}}
\hspace{-0mm}
\subfigure[SAS on PTC]{\includegraphics[width=4.3cm]{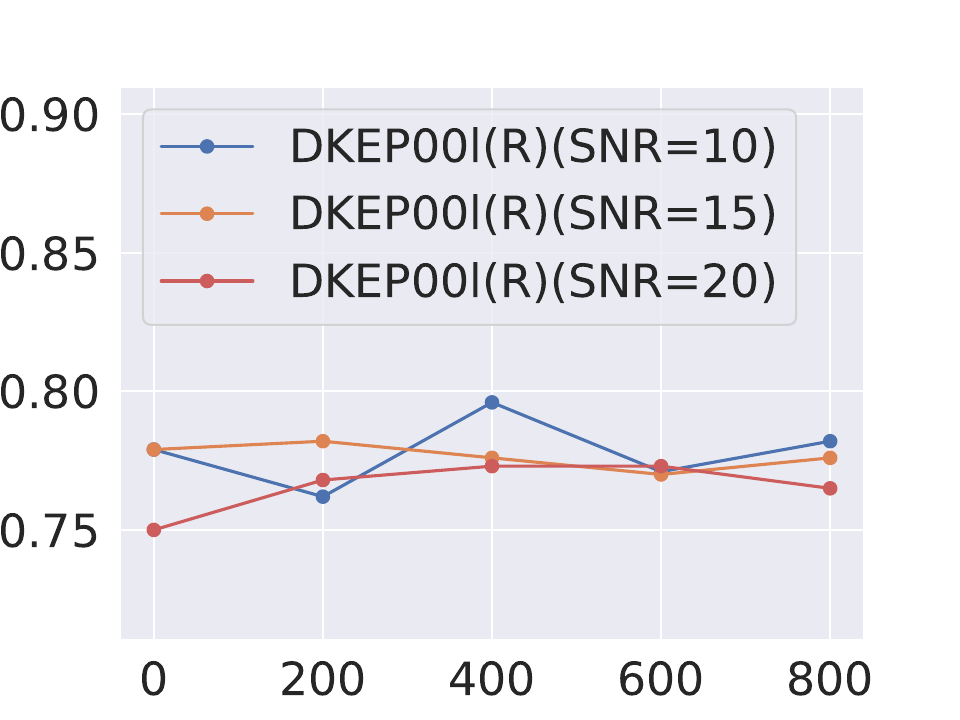}}
\hspace{-0mm}
\subfigure[SAS on PROTEINS]{\includegraphics[width=4.3cm]{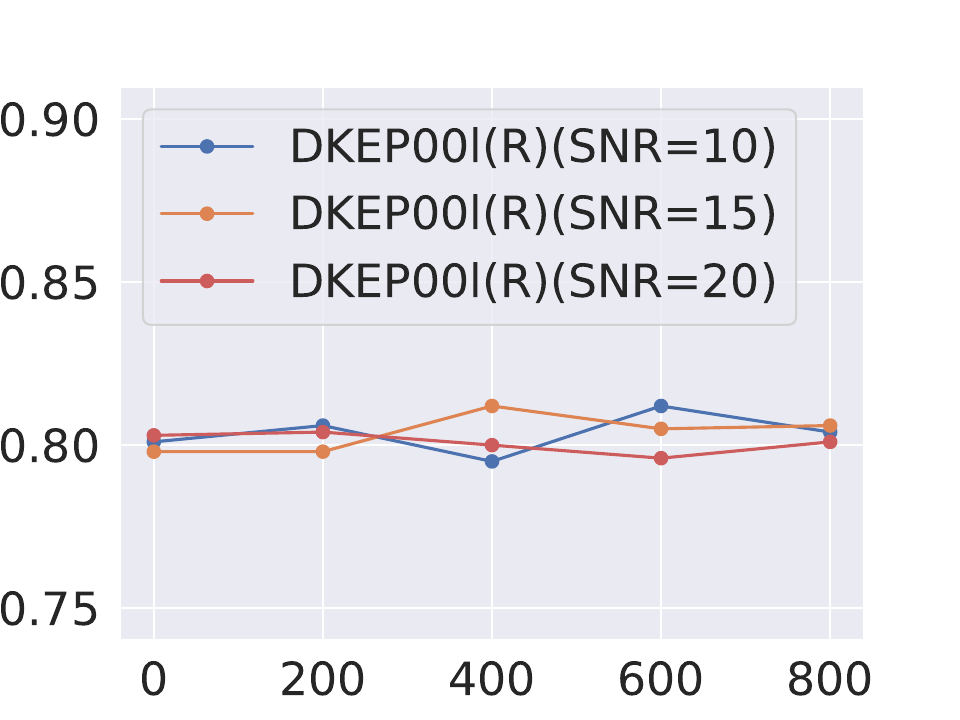}}
\hspace{-0mm}
\subfigure[SAS on RDT-B]{\includegraphics[width=4.3cm]{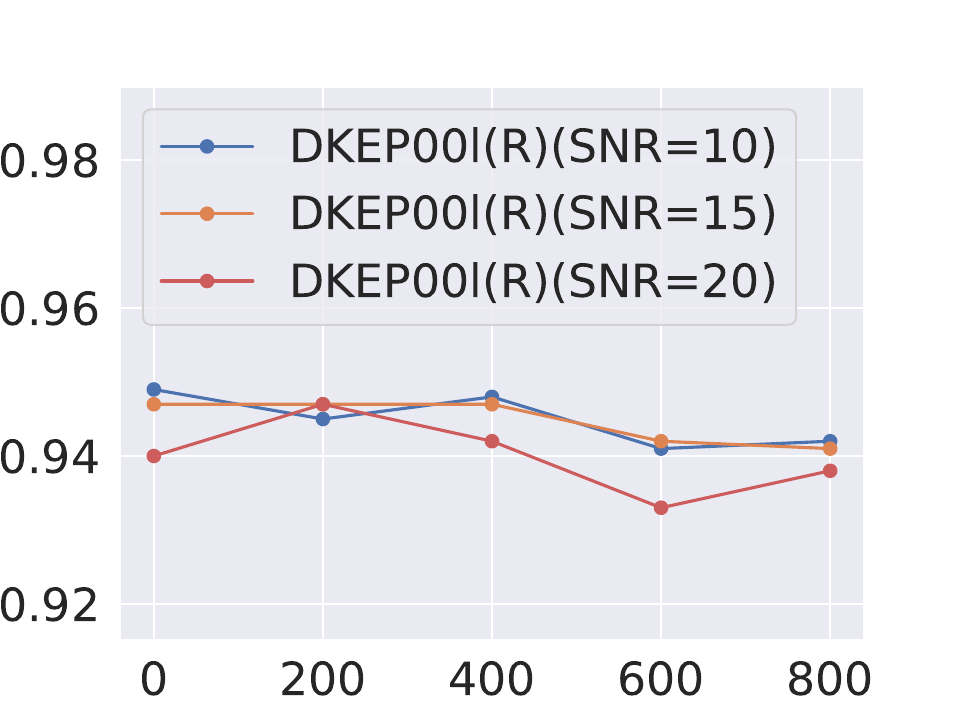}}
\hspace{-2mm}

\vspace{1mm}

\hspace{-2mm}
\subfigure[SAS on NCI1]{\includegraphics[width=4.3cm]{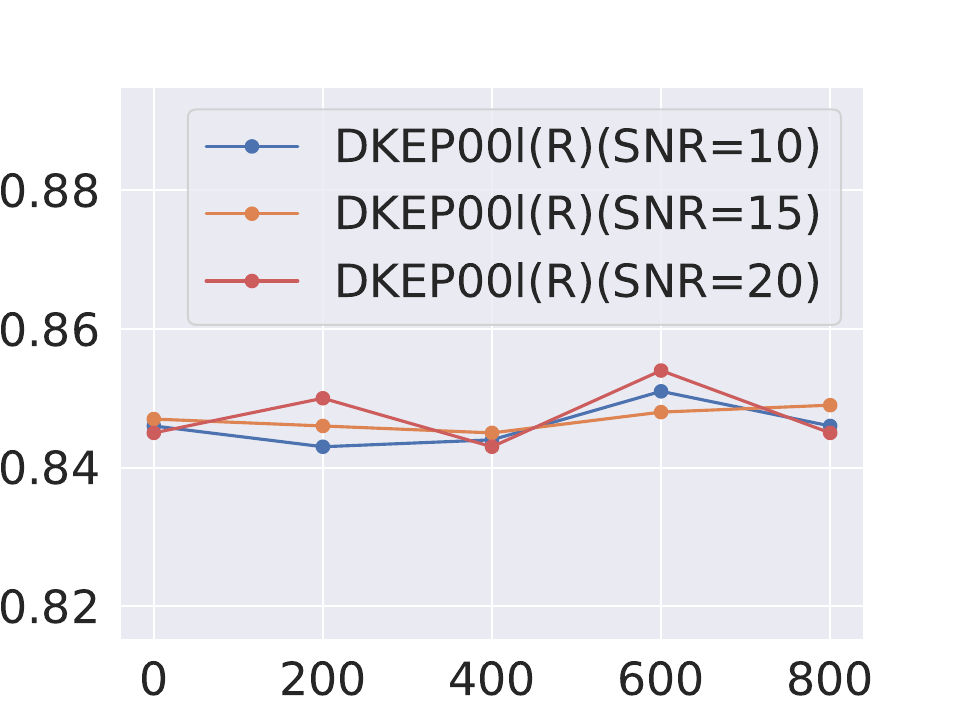}}
\hspace{-0mm}
\subfigure[SAS on IMDB-B]{\includegraphics[width=4.3cm]{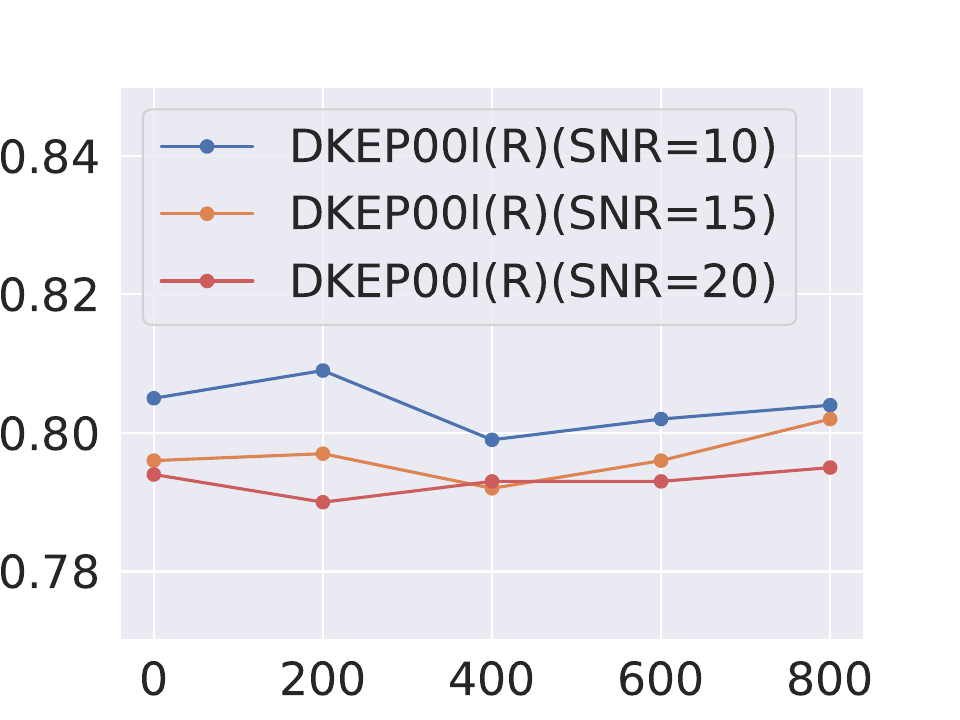}}
\hspace{-0mm}
\subfigure[SAS on IMDB-M]{\includegraphics[width=4.3cm]{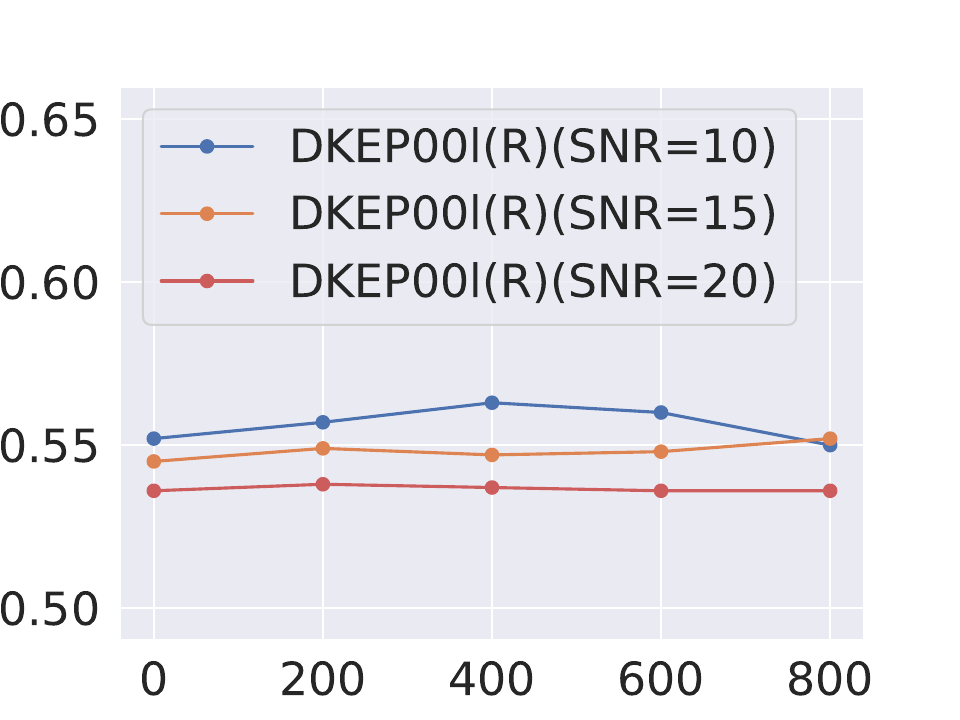}}
\hspace{-0mm}
\subfigure[SAS on MOLBBBP]{\includegraphics[width=4.3cm]{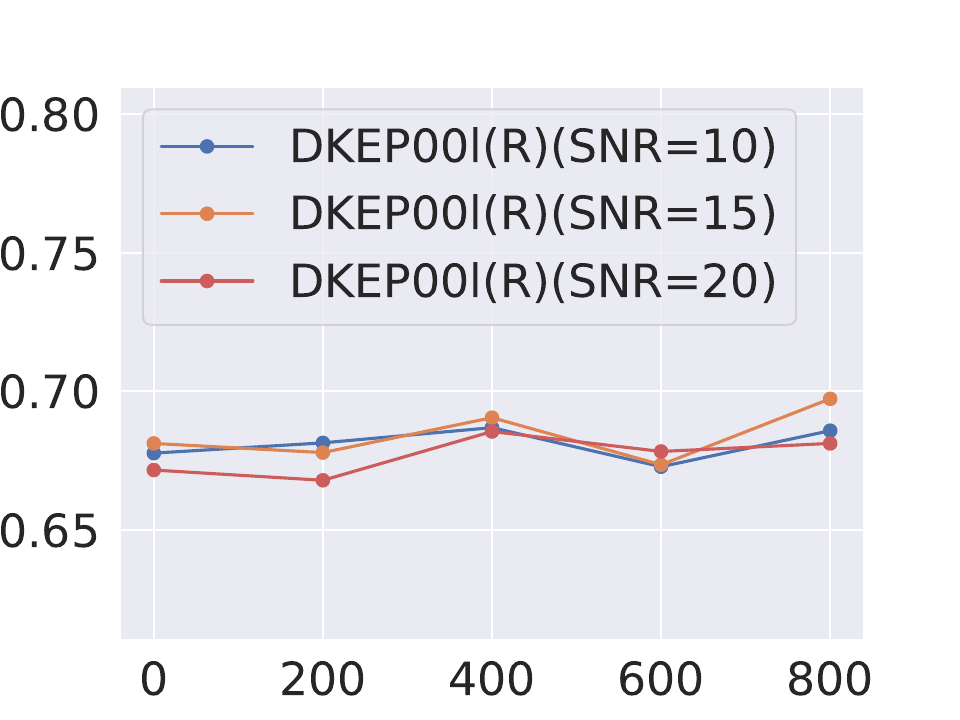}}
\hspace{-2mm}

\caption{\label{sas_ablation}
Ablation studies of SNR in DKEPool$_R$ with different representation dimensions. The abscissa refers to the representation dimension $d$, and ordinate refers to the recognition rates.
}
\vspace{-0.0cm}  
\end{figure*}

\subsection{DKE module for stepwise information extracting}

In the sub-section~\ref{sotacompare} and~\ref{gaussiancompare}, we used the DKE module as the readout operation to obtain the representation vectors of graphs, and compared them with the recent state-of-the-art methods and Gaussian-based representations. Moreover, the DKE module can be used for stepwise information extracting. In this sub-section, we applied the proposed DKE module into two hierarchical pooling frameworks, \emph{i.e.}, DiffPool~\cite{ying2018hierarchical} and MinCutPool~\cite{BianchiGA20}, for the sake of the improvement of their original performance. 
 
Table~\ref{table:hierarchical_dke} shows that two hierarchical methods that adopted DKE module improved at least 1.90\% and 1.41\% compared with their original performances on PROTEINS and MOLHIV respectively. Thus, we can conclude that our proposed DKE module can not only be used as the readout operations at the final layer but also applied into hierarchical frameworks to improve their performances. In Table~\ref{table:hierarchical_dke}, the DKE Loss $\mathcal{L}_{dke}=||\bm{z}_{pre}-\bm{z}_{post}||_2$ is designed to preserve the distribution information while pooling graph to a smaller size, where $\bm{z}_{pre}$ denotes the graph-level representation before the hierarchical pooling and $\bm{z}_{post}$ denotes the graph-level representation after hierarchical pooling. The final loss function of two methods with DKE Loss can defined as: $\mathcal{L}_{all} = \mathcal{L}_{ori} + \mathcal{L}_{dke}$, where $\mathcal{L}_{ori}$ is the original loss. 
In DiffPool and MinCutPool framework, the readout operations are the max and mean pooling respectively, and we replace original readout operations using the DKE module to achieve the final representation. The corresponding results are presented as DiffPool$_{dke}$ and MinCutPool$_{dke}$ in Table~\ref{table:hierarchical_dke}. Note that the DKE module used for these hierarchical pooling frameworks denotes its robust version with the fixed SNR of 15, and the projected matrix $\bm{W}$ is removed.

\begin{table*}
\vspace{-0.0cm}  
\caption{Ablation studies of DKEPool$_R$ with regard to iteration parameter $k$ on all nine datasets. The best results on each dataset are highlighted with \textbf{boldface.}}
\label{ablation_k_table}
\renewcommand{\arraystretch}{1.11} 
\huge
\centering 
\resizebox{\textwidth}{!}{
\begin{tabular}{cccccccccc}
\toprule
&IMDB-B 		&IMDB-M 		&MUTAG  		&PTC  	&PROTEINS  	&RDT-B  		&NCI1			&HIV		&BBBP  \\

\midrule

$k=2$
&78.5$\pm$2.8	&53.9$\pm$2.6	&95.2$\pm$3.9	&75.0$\pm$5.2   &79.1$\pm$3.6 	&93.7$\pm$1.2   &83.8$\pm$1.3	&77.4$\pm$1.4	&66.2$\pm$2.3\\

$k=4$	
&79.5$\pm$2.9	&54.3$\pm$2.1	&95.7$\pm$4.0 	&76.1$\pm$5.9 	&79.4$\pm$4.2	&93.9$\pm$1.0   &83.9$\pm$1.8	&77.9$\pm$0.9	&67.6$\pm$2.1\\

$k=5$	
&\textbf{80.2$\pm$2.3}	&\textbf{54.5$\pm$2.6} 	&\textbf{96.3$\pm$4.2}	&\textbf{77.7$\pm$3.9}	&79.6$\pm$3.9 	&\textbf{94.7$\pm$1.0}   &\textbf{84.5$\pm$1.6}	&78.1$\pm$1.7	&\textbf{68.2$\pm$2.1}\\

$k=6$	
&79.6$\pm$2.6	&54.2$\pm$1.9 	&96.2$\pm$4.7	&77.0$\pm$5.5	&\textbf{80.2$\pm$3.4}	&93.8$\pm$0.9   &84.3$\pm$1.4	&\textbf{78.8$\pm$1.2}	&67.7$\pm$1.5\\

$k=7$	
&79.4$\pm$2.1	&54.1$\pm$2.1	&95.8$\pm$4.5	&77.2$\pm$4.6	&80.0$\pm$3.6	&94.5$\pm$0.7   &84.1$\pm$1.7	&78.5$\pm$1.3	&67.0$\pm$1.9 	\\

$k=8$	
&77.7$\pm$1.8	&52.7$\pm$2.0	&95.8$\pm$4.5 	&77.3$\pm$4.9	&79.9$\pm$4.2	&94.1$\pm$1.7   &84.2$\pm$1.5	&78.2$\pm$1.4	&67.3$\pm$1.8\\
	
$k=10$	
&62.5$\pm$3.1	&34.7$\pm$2.2	&95.7$\pm$4.0 	&74.4$\pm$4.8	&63.5$\pm$4.2	&60.1$\pm$1.4   &82.5$\pm$2.0	&78.0$\pm$1.8	&66.4$\pm$2.0\\



\bottomrule

\end{tabular}
}
\vspace{-0.0cm}
\end{table*}

\subsection{Ablation studies}\label{ablation_study}

The results in Table~\ref{TUResults}, Table~\ref{OGBResults}, Table~\ref{gaussian_table}, and Table~\ref{table:hierarchical_dke} demonstrate the superior performance of our proposed pooling methods for both flat and hierarchical pooling frameworks. In this sub-section, we first investigate the ablation studies about the sensitivity of our proposed methods with regard to representation dimension and the sensitivity of the robust variant DKEPool$_R$ with regard to SNR. Then, we provide the sensitivity of the proposed DKEPool$_R$ with regard to the number of iteration $k$ in Eq.(\ref{iSQRT}). Finally, we provide the time complexity analysis of the proposed algorithm.

For the first ablation experiment, the results of MUTAG, PTC, PROTEINS, RDT-B, NCI1, IMDB-B, IMDB-M and MOLBBBP datasets have been shown in Figure~\ref{das_ablation} and Figure~\ref{sas_ablation}~(The abscissa equal to $0$ denotes that the projection matrix $\bm{W}$ is removed). Figure~\ref{das_ablation} denotes the dimensional ablation study (DAS) for our proposed modules on eight datasets respectively. Our two modules are robust to the representation dimension. Even if the projection matrix $\bm{W}$ is removed, our model can still obtain a competitive performance. Furthermore, DKEPool$_R$ usually performs better than DKEPool while their dimensions are equal, 
and the advantage is particularly obvious on the IMDB-B and IMDB-M datasets.
Figure~\ref{sas_ablation} denotes the SNR ablation study (SAS) for DKEPool$_R$ on eight datasets respectively. The DKEPool$_R$ module is more sensitive to SNR on MUTAG and PTC datasets than other several datasets. Comparing the performance with different SNR, the DKEPool$_R$ has a relatively stable performance while SNR = 15.

\begin{figure}
\vspace{-0.3cm}  
\centering
\hspace{-5mm}
\subfigure[recognition rates with regard to $k$]{\includegraphics[width=4.8cm]{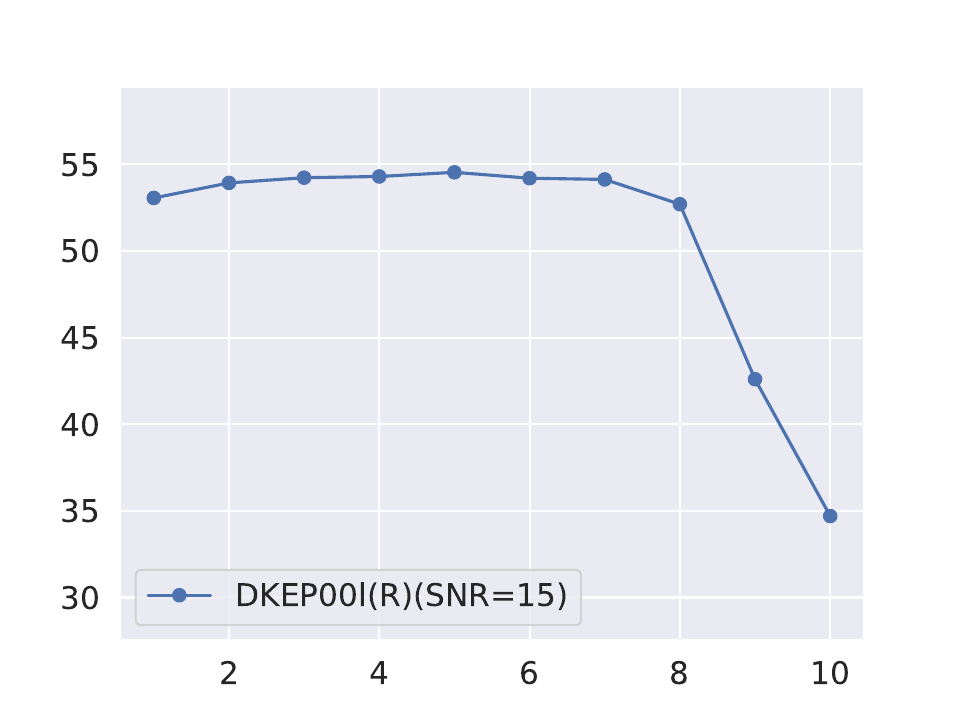}}
\hspace{-5mm}
\subfigure[tensor means with regard to $k$]{\includegraphics[width=4.8cm]{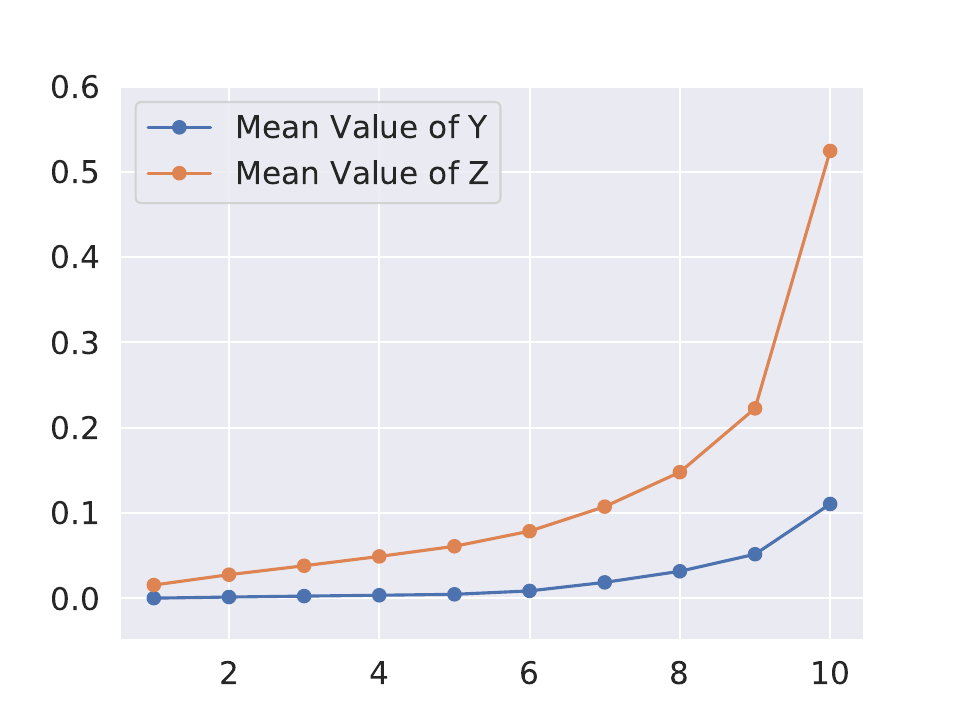}}
\hspace{-5mm}

\caption{\label{ablation_k_fig}
Ablation studies of DKEPool$_R$ with regard to iteration parameter $k$ on IMDB-M dataset. The abscissa refers to iteration parameter $k$, and ordinate refers to the recognition rates in subgraph (a) and tensor mean values in subgraph (b).
}
\end{figure}

To investigate the sensitivity of the proposed DKEPool$_R$ with regard to the number of iteration $k$ in Eq.(\ref{iSQRT}), we provide the ablation study results on all nine datasets in Table~\ref{ablation_k_table}. Here, the DKEPool$_R$ removes the projection matrix $\bm{W}$ and sets the SNR to be 15  for effective analysis. As shown in  Table~\ref{ablation_k_table}, the proposed method can achieve better performance while $k$ is between 4 and 8.
However, the performance of the DKEPool$_R$ weakens after $k=8$ on several datasets. Let's reconsider the iteration terms in Eq.(\ref{iSQRT}), the intensities in $\bm{Z}_{k}, \bm{Y}_{k}$ will be close to $f^3$ if all values in $\bm{Z}_{k-1}, \bm{Y}_{k-1} \in \mathbb{R}^{f  \times f}$ are close to 1, which means that the resulting representation vector will be perturbed by the dimension $f$ with the iteration $k$ increasing. To verify the above assumption, we plot the line charts of recognition rates and mean values of tensor $\bm{Z}_{k}$ and $\bm{Y}_{k}$ with regards to $k$ on IMDB-M dataset in Fig.~\ref{ablation_k_fig}. As shown in two subgraphs of Fig.~\ref{ablation_k_fig}, the performance decreases and the mean value increases substantially after $k=8$. Thus, we can conclude that the key information will be perturbed while the iteration $k$ increases, \emph{i.e.}, the intensities in  $\bm{Z}_{k}$ and $\bm{Y}_{k}$ are more related to the value of $f^3$ rather than $\bm{Z}_{k-1}$ and $\bm{Y}_{k-1}$ in the previous iteration. 

Finally, given the feature matrix $\bm{H} \in \mathbb{R}^{n  \times f}$ of a graph, where $n$ is the number of nodes and $f$ is the dimension of node feature, the time complexity of common readout like mean and sum operation is $\mathcal{O}(nf)$. Furthermore, the time complexity of mean centralization and covariance computation is $\mathcal{O}(nf)$ and $\mathcal{O}(nf^2)$. Thus, the final time complexity of DKEPool without affine is  $\mathcal{O}(nf^2 + f^2 + 2nf)$ where $\mathcal{O}(f^2)$ is the  time complexity of $\bm{\Sigma\mu}$. The time complexity of Eq.(\ref{iSQRT}) is $\mathcal{O}(2f^3+f^2)$, and thus the final time complexity of DKEPool$_R$ is  $\mathcal{O}(2f^3 + nf^2 + 2f^2 + 2nf)$.
\section{Conclusion}\label{Conclusion}
In this paper, we propose a practical plug-and-play module for graph-level representation learning. We first argue that distribution knowledge is crucial to graph-level downstream tasks, because the distribution space is suitable to outline the non-Euclidean geometry information of graphs. To embed the Gaussian into the linear space with Euclidean operation, we propose distribution knowledge embedding (DKEPool), a novel Gaussian-based representation module, and provide theoretical analysis to support why our DKEPool can outline distribution information. Furthermore, we introduce its robust variant based on the robust estimation of covariance in the Gaussian setting.

Extensive experiments on graph classification tasks demonstrate that the proposed DKEPool significantly and consistently outperforms the state-of-the-art methods. For the task of stepwise information extracting, we design the stepwise experiments to demonstrate that the DKE module can be applied into hierarchical pooling frameworks to improve the performance. Furthermore, there are incredible opportunities for the DKE module to be further extended to more real-world applications such as recognition tasks in computer vision because it is a general framework for representation learning.

\bibliographystyle{IEEEtran}
\bibliography{ref}

\end{document}